\documentclass[10pt,journal,compsoc]{IEEEtran}
\ifCLASSOPTIONcompsoc

  \usepackage{array}
  \usepackage{booktabs}
  \usepackage{multirow}
  \usepackage{tabularx}% 更多表格功能
  \usepackage{float}
  \usepackage{pifont}
  \usepackage{amsthm,amssymb,amsfonts}
  \usepackage{makecell}
  \usepackage{arydshln}
  \usepackage[strings]{underscore}
\else
  \usepackage{cite}

\fi
\ifCLASSINFOpdf
   \usepackage[pdftex]{graphicx}
\else
\fi
\usepackage{amsmath}
\usepackage{enumerate}
\usepackage{booktabs}
\usepackage{microtype}
\usepackage[capitalise]{cleveref}
\usepackage{color}
\usepackage{marvosym}
\usepackage{url}

\begin{document}
\title{A Survey on Knowledge-Enhanced Pre-trained Language Models}

\author{Chaoqi Zhen,
  Yanlei Shang\textsuperscript{\Letter},
  Xiangyu Liu,
  Yifei Li, 
  Yong~Chen\textsuperscript{\Letter},
  and 
  Dell Zhang,~\IEEEmembership{Senior~Member,~IEEE}
  % <-this % stops a space
\IEEEcompsocitemizethanks{
	\IEEEcompsocthanksitem Chaoqi Zhen, Yanlei Shang, Xiangyu Liu, Yifei Li, and Yong Chen are with the State Key Laboratory of Networking and Switching Technology, School of Computer Science (National Pilot Software Engineering School), Beijing University of Posts and Telecommunications, Beijing, 100876, China.
  \IEEEcompsocthanksitem Dell Zhang is with Thomson Reuters Labs, London, UK.
	\IEEEcompsocthanksitem \textsuperscript{\Letter}Yanlei Shang and Yong Chen are the corresponding authors.\protect\\E-mail: \{shangyl, yong.chen\}@bupt.edu.cn
%E-mail: A@126.com, B@bupt.edu.cn,C@bupt.edu.cn, D@hotmail.com.
}% <-this % stops an unwanted space
\thanks{Manuscript received Dec. 27, 2022; revised May **, ****.}
}

%   ^------------^------------^----Do not want these spaces!

\markboth{IEEE Transactions on Knowledge and Data Engineering,~Vol.~*, No.~*, JAN.~2023}%
{Shell \MakeLowercase{\textit{et al.}}: Bare Demo of IEEEtran.cls for Computer Society Journals}
% *** name in the headers of peer review papers. ***



\IEEEtitleabstractindextext{%
\begin{abstract}
Natural Language Processing (NLP) has been revolutionized by the use of Pre-trained Language Models (PLMs) such as BERT. 
Despite setting new records in nearly every NLP task, PLMs still face a number of challenges including poor interpretability, weak reasoning capability, and the need for a lot of expensive annotated data when applied to downstream tasks.
By integrating external knowledge into PLMs, \textit{\underline{K}nowledge-\underline{E}nhanced \underline{P}re-trained \underline{L}anguage \underline{M}odels} (KEPLMs) have the potential to overcome the above-mentioned limitations.
In this paper, we examine KEPLMs systematically through a series of studies. 
Specifically, we outline the common types and different formats of knowledge to be integrated into KEPLMs, detail the existing methods for building and evaluating KEPLMS, present the applications of KEPLMs in downstream tasks, and discuss the future research directions. 
Researchers will benefit from this survey by gaining a quick and comprehensive overview of the latest developments in this field.
\end{abstract}

}

\maketitle

\IEEEdisplaynontitleabstractindextext

% \fi
\IEEEpeerreviewmaketitle

%-----------------------------------------------------------%
\IEEEraisesectionheading{\section{Introduction}\label{sec:introduction}}

\IEEEPARstart{P}re-trained language models (PLMs) are first trained on a large dataset and then directly transferred to downstream tasks, or further fine-tuned on another small dataset for specific NLP tasks.
Early PLMs, such as Skip-Gram~\cite{b1} and GloVe~\cite{b2}, are shallow neural networks, and their word embeddings (learned from window-sized contexts) are static semantic vectors, which makes them unable to deal with the problems of polysemy in dynamic environments. 
With the development of deep learning, researchers have tried to leverage deep neural networks to boost tasks' performances with dynamic semantic embeddings. 
At first, people were still limited to the paradigm of supervised learning and thought without enough labeled data it would be difficult to unleash the potential of deep learning.  
However, with the emergence of self-supervised learning, big language models such as BERT~\cite{b3} can learn a lot of knowledge from large-scale unlabeled text data by predicting tokens that have been covered up in advance.
Thus they have made breakthrough progress in a number of downstream NLP tasks. 
Since then, many large models have started to adopt Transformer~\cite{b4} structures and self-supervised learning to solve NLP problems, and gradually PLMs have entered a phase of rapid development. 
The latest phenomenal success for PLMs is OpenAI's ChatGPT\footnote{\url{https://chat.openai.com/chat}}.

\begin{figure}[!tb]
  \centering
  \includegraphics[width=\columnwidth]{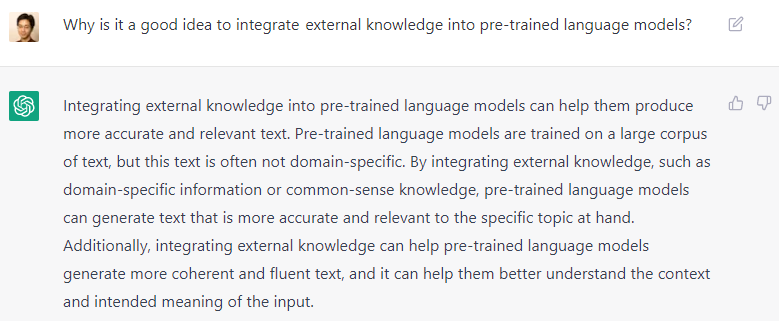}
  \caption{The benefits of integrating external knowledge into PLMs, according to ChatGPT --- one of the largest PLMs today.}
  \label{fig:ChatGPT-knowledge}
\end{figure}

As research has progressed, it has been found that PLMs still struggle with poor interpretability, weakness in robustness, and a lack of reasoning ability. 
Specifically, PLMs are widely recognized as black boxes whose decision process is opaque, thus making them difficult to interpret.
Additionally, PLMs may not be sufficiently robust as deep neural models are susceptible to adversarial examples. 
Furthermore, PLMs are also limited in their reasoning abilities because they are purely data-driven.
All these shortcomings of PLMs can be improved by incorporating external knowledge, which leads to what we call \emph{Knowledge-Enhanced Pre-trained Language Models} (KEPLMs).
Fig.~\ref{fig:ChatGPT-knowledge} shows the advantages of KEPLMs in the words of ChatGPT. 

% Many NLP researchers have studied KEPLMs and made significant contributions. 
% To name a few,
% Hernandez et al.~\cite{b5} examine various methods to integrate structured knowledge into pre-trained language models; 
% Yin et al.~\cite{b6} review PLMs regarding knowledge sources, knowledge-intensive NLP tasks, and knowledge incorporation methods; 
% Wei et al.~\cite{b7} and Yang et al.~\cite{b8} categorize PLMs from three different perspectives. 
% As time progresses, more and more methods have emerged. 
% Therefore, we conduct this survey to comprehensively introduce the latest advancements in KEPLMs for researchers.

% \begin{enumerate}[ 1.]
%     \item Present the various types and formats of knowledge being incorporated into PLMs (see \cref{sec:knowledge-sources}).
%     \item Describe the different approaches to building PLMs (see \cref{sec:Building-KEPLMs}).
%     \item Comb evaluation metrics of KEPLMs from the perspective of knowledge capacity, efficiency, and effectiveness.
%     \item Collect the typical applications (e.g., named entity recognition, question answering, and sentiment analysis) in downstream knowledge-intensive tasks.
%     \item Discuss the potential research directions (e.g., building unified KEPLMs, improving KEPLMs' interpretability and robustness) of KEPLMs in the future.
% \end{enumerate}

Although there exist a few overviews or surveys of KEPLMs~\cite{b5,b6,b7,b8}, this research field is growing and expanding rapidly with many new techniques emerged.
This survey aims to provide AI researchers the most comprehensive and up-to-date picture about the latest advancements in KEPLMs from different perspectives.

The rest of this survey are organized as follows.
\cref{sec:background} explains the background of KEPLMs.
\cref{sec:knowledge-sources} categorizes the commonly used types and formats of knowledge for KEPLMs.
\cref{sec:Building-KEPLMs} presents the different approaches to building KEPLMs. 
\cref{sec:Evaluating-KEPLMs} describes the possible performance metrics for evaluating KEPLMs. 
\cref{sec:Applying-KEPLMs} discusses the typical applications of KEPLMs in downstream knowledge-intensive NLP tasks. 
\cref{sec:challenges-future} outlines the future research directions of KEPLMs.
\cref{sec:conclusion} summarizes the contributions.

\begin{table*}[!tb]
\renewcommand{\arraystretch}{1.3}
    \centering
    \caption{Commonly Used Commonsense Knowledge}
    \label{tab:commonsense}
    % \resizebox{\columnwidth}{!}{%
  \begin{tabular}{l|p{2.2cm}|p{12.5cm}}
  \hline\hline
  \textbf{Knowledge Base} & \textbf{Domain} & \textbf{Model} \\
  \hline
  ConceptNet~\cite{b49} & open-domain & KERM~\cite{b50}, Zhang et al.~\cite{b51}, QA-GNN~\cite{b52}, GreaseLM~\cite{b53}, JointLK~\cite{b54}, GLM~\cite{b55}, KagNet~\cite{b56}, KG-BART~\cite{b57}, COMET~\cite{b58}, AMS~\cite{b59}, GRF~\cite{b60}, ExBERT~\cite{b61}, \mbox{Lauscher et al.~\cite{b62}}, \mbox{Guan et al.~\cite{b63}}, \mbox{Chang et al.~\cite{b64}}, \mbox{Yang et al.~\cite{b65}} \\
  \hline
  ATOMIC~\cite{b18} & social-interaction &  Guan et al.~\cite{b63}, Mitra et al.~\cite{b66} \\
  \hline
  $\text{ATOMIC}_{20}^{20}$~\cite{b67} & social-interaction, event-centered, physical & Hosseini et al.~\cite{b68} \\
  \hline
  ASER~\cite{b69} & eventuality & CoCoLM~\cite{b70} \\
  \hline\hline
  \end{tabular}
  % } 
\end{table*}

\begin{table}[!tb]
\renewcommand{\arraystretch}{1.3}
  \centering
  \caption{Commonly Used Domain Knowledge}
  \label{tab:domain-knowledge}
  % \resizebox{\columnwidth}{!}{%
  \begin{tabular}{l|p{5.5cm}}
    \hline\hline
    \textbf{Domain}      & \textbf{Model} \\ 
    \hline
    Science (Biomedical) & BERT-MK~\cite{b71}, UmlsBERT~\cite{b72}, \mbox{SMedBERT~\cite{b73}}, \mbox{KeBioLM~\cite{b74}}, \mbox{BioBERT~\cite{Jinhyuk-BioBert-Bioinform-2020}}, \mbox{Med-BERT~\cite{Ning-Med-BERT-Informatics-2022}} \\ 
    \hline
    Science (Other)      & SciBERT~\cite{Iz-SciBert-EMNLP-2019}, MatBERT~\cite{Walker-MatBert-posted-2021}\\ 
    \hline
    E-commerce & E-BERT~\cite{b75}, K-AID~\cite{b76}, K-PLUG~\cite{b77} \\ 
    \hline
    Law  & Legal-BERT~\cite{Ilias-LegalBert-EMNLP-2020,Lucia-when-ICAIL-2021}, Lawformer~\cite{Chaojun-LawFormer-AIOpen-2021} \\ 
    \hline
    Sentiment  & KET~\cite{b78}, SKEP~\cite{b79}, REMOTE~\cite{b80} \\ 
    \hline
    Programming    & GraphCodeBERT~\cite{Daya-GraphCodeBert-ICLR-2021} \\
    \hline\hline
  \end{tabular}
  % }
\end{table}
 
\section{Background}
\label{sec:background}

In this section, we first introduce the concept of PLMs, and then talk about the recent trend of combining PLMs and knowledge.

\subsection{Pre-trained Language Models}

In 2013, word2vec~\cite{b9} opened the era of pre-trained language models. 
First-generation PLMs such as Skip-Gram~\cite{b1} and GloVe~\cite{b2} aim to get good word embeddings for downstream tasks directly, and their model architectures are typically shallow neural networks to allow for computational efficiency.~\cite{b10}.
Second-generation PLMs, e.g., LSTM~\cite{b13} based CoVe~\cite{b11} and ELMo~\cite{b12} as well as Transformer~\cite{b4} based BERT~\cite{b3} and GPT~\cite{b14} focus on learning word embeddings in dynamic contexts.
During that period, Transformers became most successful in almost all downstream NLP tasks and brought significant changes to the NLP field.  
Today, PLMs generally refer to models based on the Transformer architecture, under the pre-train-then-fine-tune paradigm. 
The representative PLMs include GPT~\cite{b14} (an auto-regressive language model based on Transformer Decoder), BERT~\cite{b3} (an auto-encoding language model based on Transformer Encoder), and BART~\cite{b15} (a sequence-to-sequence model based on both Transformer Encoder and Decoder).
Very recently, prompt learning, a new paradigm in NLP, is getting more and more popular~\cite{b16}. 
It can help to make better use of knowledge in PLMs and hence empower PLMs with the ability to perform few-shot or even zero-shot learning for challenging scenarios with little or none labeled data.

\subsection{Knowledge and PLMs}

There are two lines of research on the interaction between knowledge and PLMs: one is to use PLMs as Knowledge Bases (KBs), and the other is to enhance PLMs with knowledge.
In this paper, we focus on the latter.

\subsubsection{Using PLMs as Knowledge Bases}
\label{sec:plms-as-kbs}

KBs (e.g., Wikidata~\cite{b17} and ATOMIC~\cite{b18}) store entities and their relationships, usually in the form of relation triplets. 
PLMs are considered as a possible alternative to structured KBs, which has attracted many researchers. 
Beginning with LAMA~\cite{b19}, many researchers explored whether PLMs could serve as structured KBs. 
Wan et al.~\cite{b20} explored how to construct KBs using pre-trained language models automatically. 
Heinzerling et al.~\cite{b21} investigated the relationship between accuracy and memory capacity of neural networks, arguing that PLMs can be used as KBs. 
Safavi et al.~\cite{b22} argued that relational KBs represent knowledge with high accuracy but lack flexibility. 
%Training language models to internalize and express relational knowledge in the textual form can partially break the limitations of KBs, making LM and KB complement each other. 
In contrast, Razniewski et al.~\cite{b23} dived into the strengths and limitations of both PLMs and KBs. 
They believed that KBs with explicit knowledge could not be completely replaced by PLMs with latent knowledge. 
Wang et al.~\cite{b24} found that closed-book question answering is still a challenge for generative models, and therefore generative models are not suitable to serve as KBs. 
AlKhamissi et al.~\cite{b25} argued that there are five aspects at which a PLM needs to excel to qualify as a KB and found that three of them (i.e. consistency, reasoning, and interpretability) are better obtained in KBs than in PLMs.
%There is a long way to go before PLMs can act as KBs.

\subsubsection{Enhancing PLMs with Knowledge}

On the other way around, we can use knowledge to improve or extend PLMs. 
In many knowledge-intensive downstream tasks, taking question-answering tasks as an example, the amount of knowledge learned by the pre-trained language model can be increased by adding parameters; however, it is far less effective than directly integrating knowledge~\cite{b26}. 
Therefore, it is necessary to inject knowledge into PLMs to obtain better performance.

Methods such as ERNIE~\cite{b38}, KnowBert~\cite{b95}, K-BERT~\cite{b45} are early attempts at incorporating knowledge into PLMs, and they have achieved great success especially on knowledge-intensive NLP tasks. 
Many subsequent models were inspired by them and improved upon them. 
Nowadays, more and more KEPLMs are emerging, integrating different kinds of knowledge in different ways and dealing with a variety of NLP tasks. 
In what follows, we will present a comprehensive overview of KEPLMs.

\section{Knowledge Sources for KEPLMs}
\label{sec:knowledge-sources}

In this section, we elaborate on the common types and formats of knowledge that are incorporated into PLMs.

\subsection{Types of Knowledge}
\label{sec:knowledge-types}

There are five types of knowledge that are often integrated into PLMs: linguistic knowledge, semantic knowledge, commonsense knowledge, encyclopedic knowledge, and domain knowledge.

\subsubsection{Linguistic Knowledge}

% Linguistic Knowledge mainly includes multilingual knowledge, part-of-speech tags, and syntactic knowledge.

\textit{Part-of-Speech Tags.} 
Commonly used part-of-speech tags include pronouns, verbs, nouns, pre-positions, conjunctions, adverbs, and adjectives. 
The could help the understanding of natural language text data, e.g., for sentiment analysis. 
% Every word must be marked correctly. 
% For example, the term ``great'' in the sentence ``I require a great deal of help'' should be tagged as negative. 
SentiLARE~\cite{b28} exploits part-of-speech tags to promote sentiment analysis.

\textit{Syntactic Structures.} 
Models acquire the structure of sentences via syntactic parsing, mainly including constituency and dependency. 
Syntax-BERT~\cite{b29} uses syntax-related masks to incorporate information from constituency and dependency trees.
K-Adapter~\cite{b30} integrates dependency parsing information, which has improved the performance of the dependency relation prediction task.

\textit{Cross-lingual Transferability.} 
Sometimes PLMs could obtain cross-lingual transferability through learning from multilingual corpora. 
For example, as demonstrated by XLM-K~\cite{b27}, the linguistic knowledge about one language might help the processing of another language.

\subsubsection{Semantic Knowledge}
Semantic knowledge aims to help models catch the meaning of texts. For example,
KT-NET~\cite{b31}, SenseBERT~\cite{b32}, and LIBERT~\cite{b33} introduce semantic knowledge from WordNet~\cite{b34} and perform well on machine reading comprehension, word sense disambiguation, and lexical simplification, respectively.

Basu et al.~\cite{b35} converted the syntax tree of the text to its corresponding semantic meanings with the help of VerbNet~\cite{b36} so that the model could understand the text. 
SemBERT~\cite{b37} incorporates semantic knowledge from semantic role labeling for better reading comprehension and language inference.

\subsubsection{Commonsense Knowledge}

Commonsense knowledge is the routine knowledge people have of their everyday world and activities~\cite{b48}. Commonly used knowledge bases are listed in Table~\ref{tab:commonsense}.

Commonsense knowledge is represented as triples in KBs, where head and tail entities are more often phrases than just words, which are different from encyclopedic knowledge. 
For example, a commonsense triple is like \textit{(having no food, CauseDesire, go to a store)}, while an encyclopedic triple is like \textit{(China, capital, Beijing)}.

As shown in Table 1, ConceptNet~\cite{b49} is the most widely used open-domain commonsense knowledge graph containing 34 relations, such as RelatedTo, IsA, Causes, etc. 
It's helpful in commonsense question answering, commonsense validation, and commonsense story generation.

ATOMIC~\cite{b18} focuses on inferential knowledge organized by \textit{“if-then”} structure, e.g., “\textit{if X pays Y a compliment, then Y will likely return the compliment}”, covering \textit{causes} vs. \textit{effects}, \textit{agents} vs. \textit{themes}, \textit{voluntary} vs. \textit{involuntary events}, and \textit{actions} vs. \textit{mental states}, which is helpful for commonsense reasoning. 
It can be employed in commonsense question generation and commonsense question-answering tasks, such as Guan et al.~\cite{b63} and Mitra et al.~\cite{b66}.

$\text{ATOMIC}_{20}^{20}$~\cite{b67} covers more accurate and diverse commonsense knowledge than the aforementioned commonsense knowledge sources, including three categories of social interaction, physical and event-centered. 
Hosseini et al.~\cite{b68} convert triples (in $\text{ATOMIC}_{20}^{20}$~\cite{b67}) into natural language sentences for pre-training, which improves the performance on causal pair classification and commonsense of answering questions tasks.

ASER~\cite{b69} is a large-scale eventuality knowledge graph, with events as nodes and discourse relations as edges. 
It provides more complicated commonsense knowledge, such as the cause-effect relation between \textit{``Jim yells at Bob'' }and \textit{``Bob is upset''}.

\subsubsection{Encyclopedic Knowledge}

Encyclopedic knowledge covers widespread information in the open domain, in the form of texts or triples. 
Wikipedia\footnote{{https://www.wikipedia.org/}} is a multilingual encyclopedia that is unstructured. 
BERT utilizes Wikipedia as pre-training data to learn contextual representations; while other methods usually leverage encyclopedic knowledge via knowledge triples.

Wikidata~\cite{b17} is the most widely used knowledge graph when incorporating encyclopedic knowledge. 
KEPLMs such as K-Adapter~\cite{b29}, ERNIE~\cite{b38}, KgPLM~\cite{b39}, and ERICA~\cite{b40}, use Wikidata~\cite{b17} as knowledge sources. 
Other commonly used English encyclopedic knowledge graphs include Freebase~\cite{b41}, DBpedia~\cite{b42}, and NELL~\cite{b43}. 
CN-DBpedia~\cite{b44} is a widely used Chinese encyclopedic knowledge graph. 
KEPLMs designed for Chinese downstream tasks, such as K-BERT~\cite{b45}, use CN-DBpedia as knowledge sources.

Wikidata5M is a large-scale knowledge graph proposed by KEPLER~\cite{b46} that contains high-quality descriptions of entities and relations in addition to triples. KEPLER~\cite{b46} used these descriptions to initialize knowledge embeddings, and CoLAKE~\cite{b47} also adopted this approach.

\subsubsection{Domain Knowledge}

In contrast to encyclopedic knowledge, domain knowledge is knowledge of a specific, specialized field discipline, such as biomedical, e-commerce, and sentiment, which are explored a lot, as shown in Table~\ref{tab:domain-knowledge}.

Biomedical knowledge is usually represented as triples containing symptoms or diseases as head or tail entities, e.g., \textit{(bacterial pneumonia, with associated morphology, inflammation)}. 
E-commerce knowledge is formed with product names, while their descriptions are represented by a set of phrases. 
For example, the product ``iPhone XS'' is described as ``iOS; 4G signal; T-Mobile service; OLED screen; ...''. 
Sentiment knowledge could be represented in many ways, including sentiment words, word polarity, etc.

\subsection{Formats of Knowledge}

There are four formats of knowledge that are often incorporated into PLMs, i.e., entity lexicon, knowledge graph, plain text, and labeled images. 
% Methods choose the appropriate form depending on the problem to be solved.

\subsubsection{Entity Lexicon}

To incorporate knowledge through entities, we need to integrate knowledge embeddings of entities into aligned token embeddings. 
Existing models propose two methods to obtain initial entity embeddings, obtained through traditional knowledge embedding algorithms, such as TransE used by ERNIE~\cite{b38} and CokeBERT~\cite{b83}, or through encoding entity descriptions, such as KEPLER~\cite{b46}.

The first approach can fuse the information of neighboring nodes of entities in the knowledge base. But it must face the challenge of heterogeneous embedding space because the embedding vector space of words in the text and entities in KG is inconsistent. The second approach fuses the information in the same embedding space, but the entity embedding may not fully express the meaning of the entities.

It's simple and intuitive to inject knowledge in the form of entity embeddings. 
However, entity embeddings need to be retrained when the knowledge graph is updated, and the model parameters have to be retrained if injected in the pre-training phase.

\subsubsection{Knowledge Graph} 

\emph{\textbf{Triples}}. 
Knowledge stored in the knowledge graphs is commonly in the form of semantic (RDF) triples. 
To incorporate triples, we can append triples to the proper position in the text, such as K-BERT~\cite{b45}, ERNIE 3.0~\cite{b84}, Zhang et al.~\cite{b51}, and Bian et al.~\cite{b85}, or integrate their embeddings into text embeddings, such as Liu et al.~\cite{b86}. 
More details are in \cref{sec:explicit-incorporation}.

\emph{\textbf{Subgraphs}}. 
Knowledge subgraphs are part of knowledge graphs that take entities as nodes and relationships as edges. 
KEPLMs such as QA-GNN~\cite{b52}, GreaseLM~\cite{b53}, KG-BART~\cite{b57}, and KALA~\cite{b87} incorporated knowledge in the form of knowledge subgraphs, which are described in detail in \cref{sec:adding-knowledge-fusion-module}.

\subsubsection{Plain Text} 

To integrate knowledge in texts, we can convert knowledge triples to sentences as the pre-training corpus or add related entity definitions to texts. 
Commonsense knowledge triples are suitable to be converted into sentences. 
or instance, Hosseini et al.~\cite{b68} convert the triples \textit{(PersonX accidentally fell, xEffect, PersonX breaks an arm)} in $\text{ATOMIC}_{20}^{20}$ into the sentence ``Tracy accidentally fell. As a result, Tracy breaks an arm.'', and fed them to the model for continually pre-training. 
This method is suitable for commonsense knowledge bases because the triples in them are usually phrases and only need to add conjunctions to obtain sentences. 

Dict-BERT~\cite{b88} append the definitions of the rare words to the end of the text as model input, which facilitates the model's understanding and learning of rare words, but does not apply to polysemous words.

\subsubsection{Captioned Images} 

% This method focuses on the learning of visual knowledge. 
% VALM~\cite{b81} and Vokenization~\cite{b82} both incorporate visual knowledge in the form of images associated with corresponding text captions.

% \subsubsection{Visual Knowledge}

\begin{figure}[!tb]
  \centering
  \includegraphics[width=0.7\columnwidth]{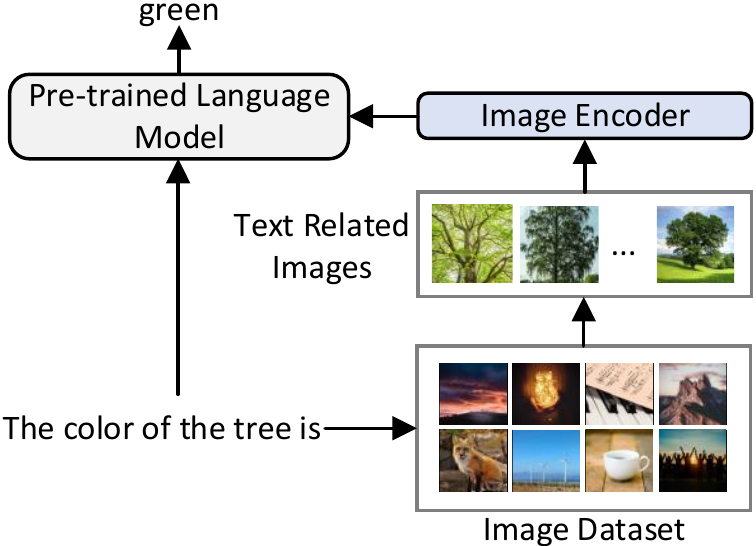}
  \caption{Incorporating visual knowledge from captioned images into PLMs.}
  \label{figure0}
\end{figure}

Unlike the knowledge presented above in the form of texts, visual knowledge refers to knowledge observed through the eyes, including the shape, size, and color of objects, which may not be mentioned in the text and need to be learned through images. 
To incorporate visual knowledge, models can first retrieve context-related images, encode them, and then integrate the embeddings of images to text embeddings, such as VALM~\cite{b81}, as shown in Fig.~\ref{figure0}.
Visual knowledge can also be incorporated through text-image alignment pre-training objectives, such as Vokenization~\cite{b82}.

%-----------------------------------------------------------%
\section{Building KEPLMs}
\label{sec:Building-KEPLMs}

When we construct KEPLMs, external knowledge could be incorporated into PLMs \emph{implicitly} and/or \emph{explicitly}.

\subsection{Implicit Incorporation of Knowledge}
\label{sec:implicit-incorporation}

\subsubsection{Knowledge-Guided Masking Strategies}

\begin{figure}[!tb]
\centering
\includegraphics[width=1.0\columnwidth]{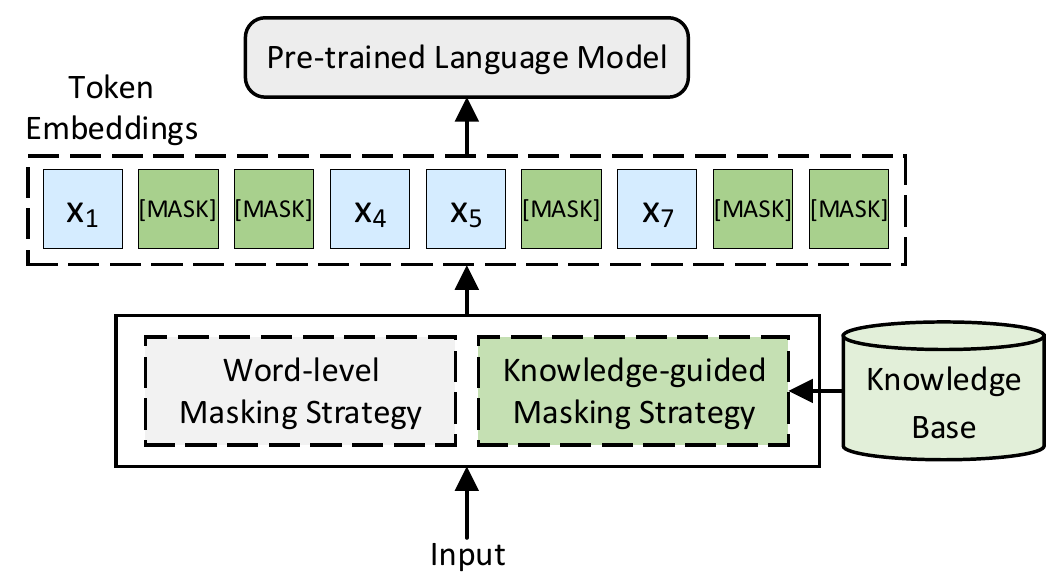}
\caption{Using knowledge-guided masking strategies to build KEPLMs.}
\label{figure1}
\end{figure}

PLMs represented by BERT generally use unstructured text documents from Wikipedia etc. as the corpus for pre-training. 
The unstructured text data contain rich contextual semantic information from which BERT could learn the contextual knowledge of words through Masked Language Modelling (MLM). 
However, entities and phrases in the text that also contain valuable information have been ignored. 
By employing a knowledge-guided masking strategy beyond the level of individual words, PLMs are able to incorporate the knowledge about entities and phrases etc., as shown in \cref{figure1}.

ERNIE~\cite{b89} adds entity-level and phrase-level masking strategies to BERT, and thus guides the pre-training of BERT to incorporate the entity and phrase information from text. 
SKEP~\cite{b79} proposes to mask not entities or phrases but sentiment words so as to inject sentiment knowledge into text representations. 

\begin{figure}[!tb]
\centering
\includegraphics[width=1.01\columnwidth]{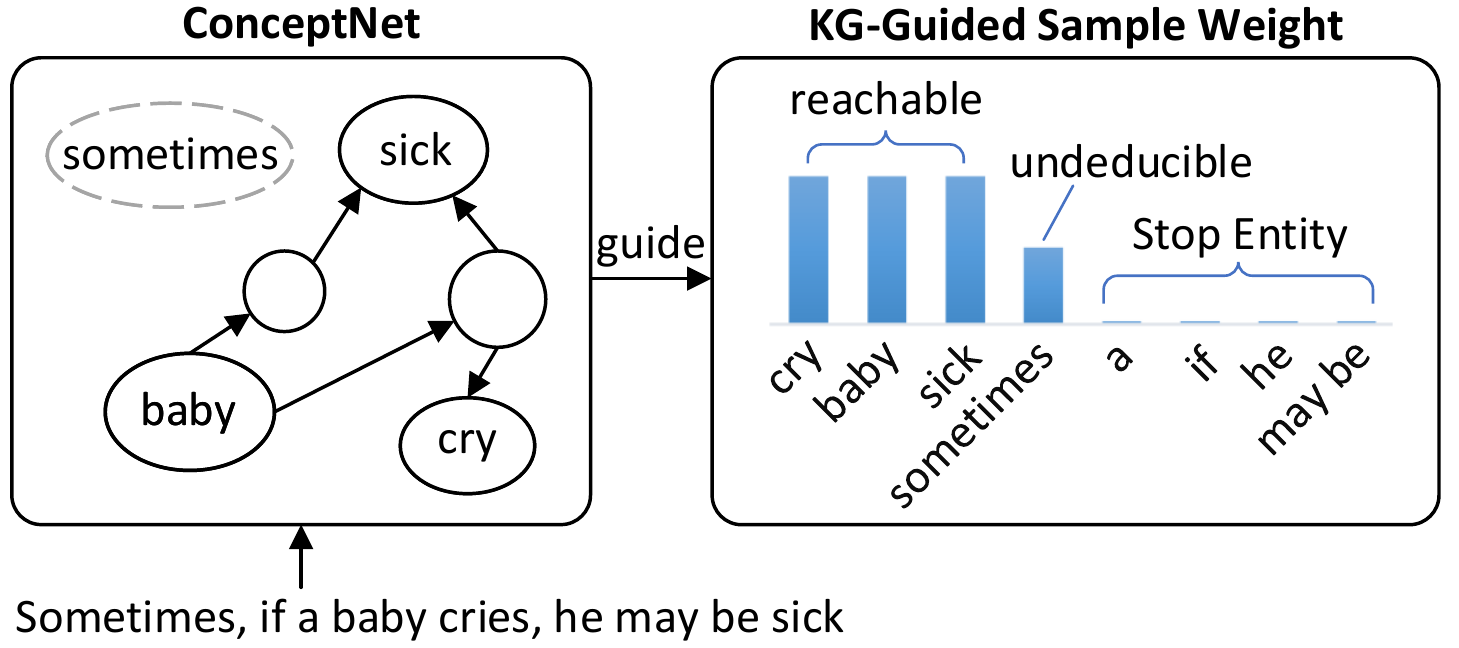}
\caption{GLM's knowledge-graph informed sampling of entities for masking.}
\label{figure2}
\end{figure}

Different from the simple random selection of entities or phrases for masking (as in ERNIE), GLM~\cite{b55} uses knowledge-graph informed sampling that assigns higher weights to more important entities.
Specifically, GLM's masking strategy would mask a general word $20\%$ of the time and an entity $80\%$ of the time. 
When GLM needs to mask an entity, those entities which can reach other entities in the sentence within a specific number of hops in ConceptNet~\cite{b49} are considered more critical and given higher probabilities to be chosen.
In this way, GLM can nudge the construction of KEPLMs towards more critical entities in the knowledge graph.
As illustrated in \cref{figure2}, assuming that among the four entities in the given sentence, three of them ``sick'', ``baby'' and ``cry'' could be reached within a specific number of hops in ConceptNet while the other one ``sometimes'' could not, GLM would give the former more chances than the latter to be masked for pre-training when entity-level masking is activated ; the other non-entity words in the sentence would be sampled only when word-level masking is activated. 
% This way, the original word-level mask for learning contextual semantics is preserved, while the model is guided to learn important entities in the sentence.

\begin{figure}[!tb]
\centering
\includegraphics[width=0.8\columnwidth]{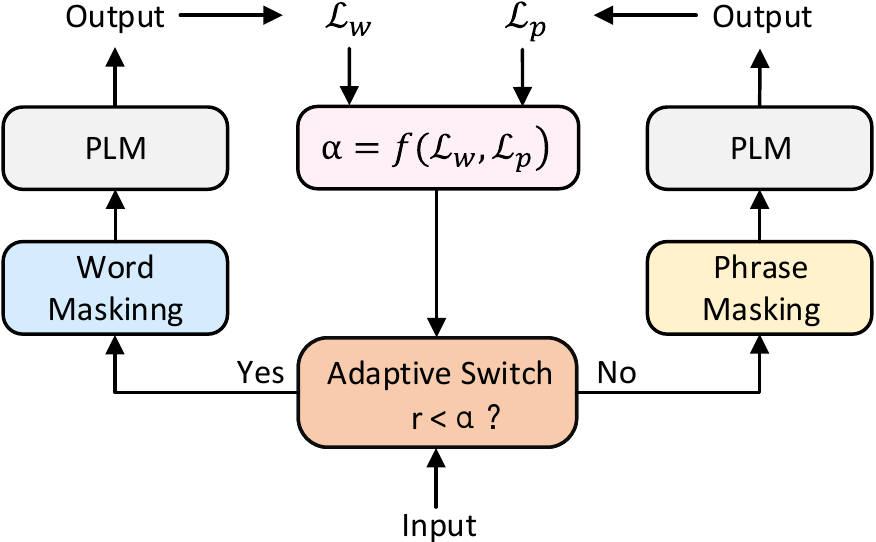}
\caption{E-BERT's adaptive hybrid masking strategy.}
\label{figure3}
\end{figure}

Instead of using a predefined probability to choose between the two modes of masking (as in GLM), E-BERT~\cite{b75} proposes an adaptive hybrid masking strategy that allows the model to switch between word-level and phrase-level masking in an adaptive fashion during its pre-training. 
As illustrated in \cref{figure3}, E-BERT~\cite{b75} enters the mode of word-level masking when $r<\alpha^t$ and the mode of phrase-masking otherwise, where $r$ is a randomly generated number in each iteration. 
The loss functions $\mathcal{L}_w$ and $\mathcal{L}_p $ of the two modes in each iteration are used to track the fitting progress of the learned word-level information and the learned phrase-level information, represented by $\eta_{w\ }^t $ and $ {\ \eta}_{p\ }^t$, respectively. 
The relative importance of word-level masking with respect to phrase-level masking, $r^t $, is used further to calculate $\alpha^{t+1}$ as in \cref{eq1}, so that the mode with higher loss in the current iteration is more likely to be selected in the next iteration. 
\begin{equation}
\label{eq1}
    \begin{aligned}
    &\eta_w^t=\mathrm{\Delta}_w^{t,t-1}/\mathrm{\Delta}_w^{t,1}=\left[\mathcal{L}_w^{t-1}-\mathcal{L}_w^t\right]_+\ /\ (\mathcal{L}_w^1-\mathcal{L}_w^t),\\
    &\eta_p^t=\mathrm{\Delta}_p^{t,t-1}/\mathrm{\Delta}_p^{t,1}=\left[\mathcal{L}_p^{t-1}-\mathcal{L}_p^t\right]_+\ /\ (\mathcal{L}_p^1-\mathcal{L}_p^t),\\
    &r^t=\eta_w^{t+1}/\eta_p^{t+1},\\
    &\alpha^{t+1}=tanh\left(r^t\right).
    \end{aligned}
\end{equation}
Thus, E-BERT can switch between the two modes of masking adaptively and strike a balance between them.

\subsubsection{Knowledge-Related Pre-training Tasks}

\begin{figure}[!tb]
\centering
\includegraphics[width=0.7\columnwidth]{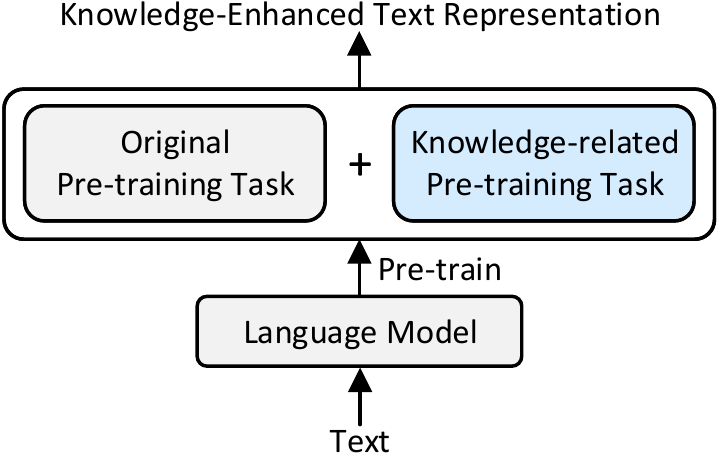}
\caption{Using knowledge-related pre-training tasks to build KEPLMs.}
\label{figure4}
\end{figure}

Some methods for building KEPLMs incorporate knowledge implicitly by adding knowledge-related pre-training tasks, as shown in \cref{figure4}.

For example, KALM~\cite{b90} enriches the input sequence with entity signals and then adds an entity prediction task to the pre-training objective in order to help the model learn entity information better. 
KEPLER~\cite{b46} adds the knowledge embedding pre-training task which shares a Transformer Encoder with the MLM, obtaining text-enhanced knowledge embeddings and knowledge-enhanced PLMs simultaneously. 
Vokenization~\cite{b82} proposes the concept of voken (visualized token), i.e., token-related images; it adds a voken classification task that predicts the image corresponding to each token so as to enhance the PLMs with visual knowledge which has been shown to help some downstream NLP tasks.

\subsection{Explicit Incorporation of Knowledge}
\label{sec:explicit-incorporation}

\begin{figure*}[!tb]
  \centering
  \includegraphics[width=1.6\columnwidth]{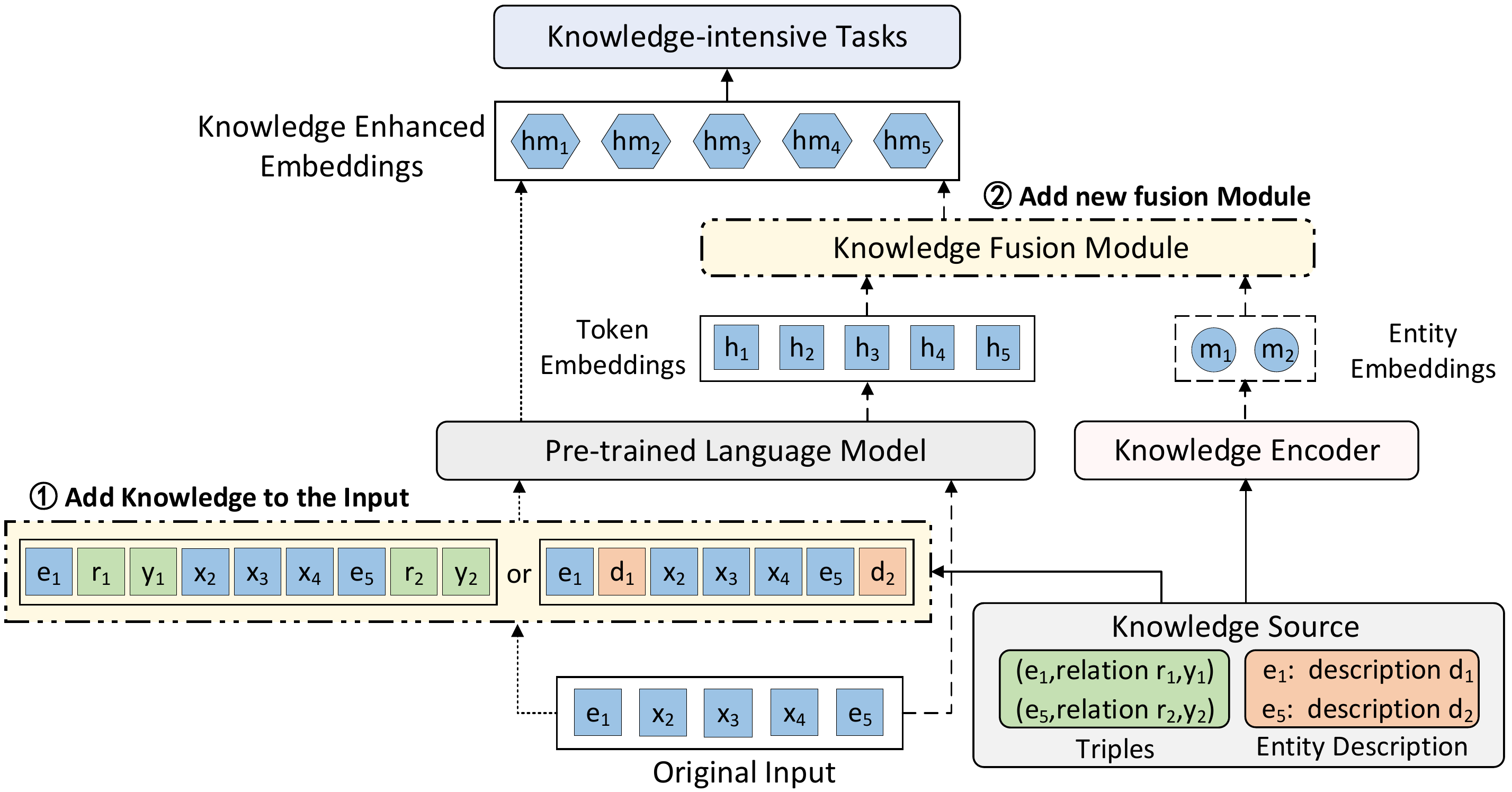}
  \caption{Explicit incorporation of knowledge into PLMs via modifying the model input or adding knowledge fusion modules.}
  \label{figure5}
\end{figure*}

There are mainly three ways for PLMs to incorporate external knowledge explicitly: modifying the model input, adding knowledge fusion modules, and utilizing external memory. 
The first two approaches insert relevant knowledge into PLMs, in the form of either additional input for the model or additional components in the model, as shown in \cref{figure5} \ding{172} and \ding{173}. 
The third approach keeps the text and knowledge spaces independent which can facilitate knowledge updates.

\subsubsection{Modifying the Model Input}
\label{sec:modifying-the-model-input}

Some KEPLMs insert relevant knowledge triples or entity descriptions into the input for the model during its pre-training.

There exist a few different ways to incorporate knowledge in the form of triples.
ERNIE 3.0~\cite{b84} prepends related triples to the sentences as the expanded model input. 
K-BERT~\cite{b45} injects relevant triples into each sentence to generate a sentence tree for model input. 
To be specific, if the input sentence has an entity ``apple'', K-BERT~\cite{b45} will find the triples whose head entity is ``apple'' in the knowledge graph and then append the relation and tail entity of these triples to ``apple'' to generate a new sentence tree. 
A visible matrix is created to control the level of knowledge noise. 
Zhang et al.~\cite{b51} improve the visible matrix of K-BERT to further minimize the introduction of knowledge noise. 
CoLAKE~\cite{b47} also introduces triples to the input text, treats the text as a fully connected word graph, and integrates knowledge to form a word-knowledge graph. 
It takes inspiration from K-BERT and makes some improvements in the reduction of knowledge noise. 
For the question answering task, Bian et al.~\cite{b85} convert multiple question-related knowledge triples into text according to predefined templates and feed them into the model together with the question and alternative answers for training, which obtains excellent performance on commonsense question answering.
For all the above methods that insert knowledge triples to the model input, the introduction of external knowledge may damage the original sentence structure, and therefore we must try to reduce knowledge noise in this process. 

There are also a few different ways to incorporate knowledge in the form of entities.
Dict-BERT~\cite{b88} obtains the definitions of rare words in a sentence from Wiktionary~\cite{b91} and appends them to the end of the sentence. 
Similarly, DKPLM~\cite{b92} focuses on long-tail entities and uses pseudo token representations from relevant triples to replace their embeddings.
Unlike the above methods, WKLM~\cite{b93} replaces entities in the text with other entities of the same type, which are then fed into the model. 
Then the model is asked to determine which entities in the sentence are correct and which are replaced. 
This method does not modify the model, only the input data during its pre-training. 
A few high-performance KEPLMs constructed using this method are described in detail below.

\begin{figure*}[!tb]
\centering
\includegraphics[width=1.5\columnwidth]{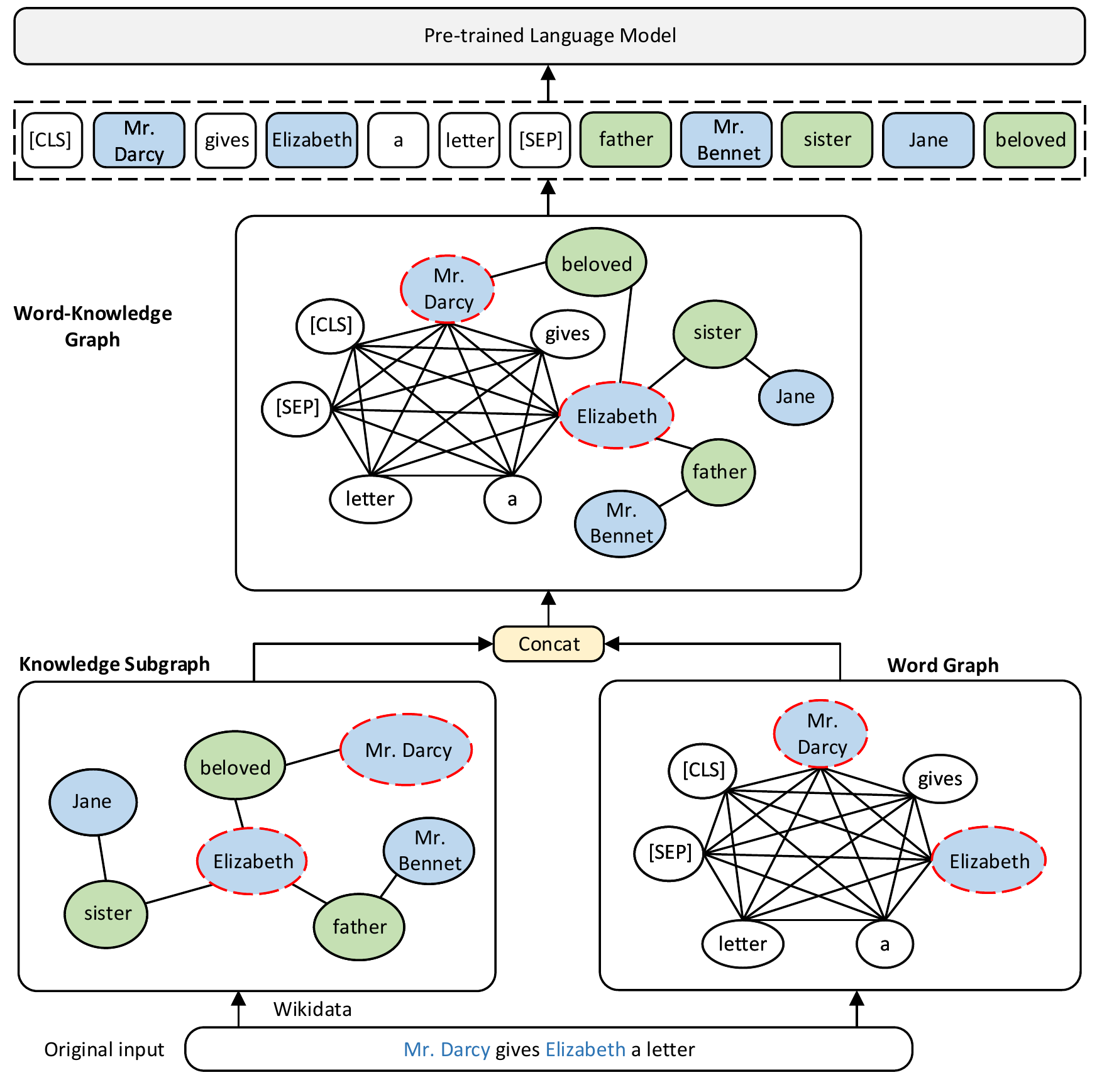}
\caption{Adding knowledge triples into the model input.}
\label{figure6}
\end{figure*}

CoLAKE~\cite{b47} modifies the model input to incorporate knowledge of entities, as shown in \cref{figure6}.
Specifically, CoLAKE regards each input sentence as a fully connected graph. 
It takes the entity in the input sentence as the anchor node and introduces a subgraph (composed of triples with that anchor node as the head entity in the knowledge graph) to obtain the word knowledge graph. 
Then the newly added nodes from the word knowledge graph are appended behind the original input text and fed into the PLMs together for pre-training. 
CoLAKE distinguishes the node types in the newly obtained input statement and initializes different nodes differently. 
These nodes include word nodes, entity nodes, and relation nodes. 
CoLAKE achieves a 5.2\% improvement on the relation classification task in comparison to BERT without knowledge integration.

\begin{figure}[!tb] 
\centering
\includegraphics[width=1.01\columnwidth]{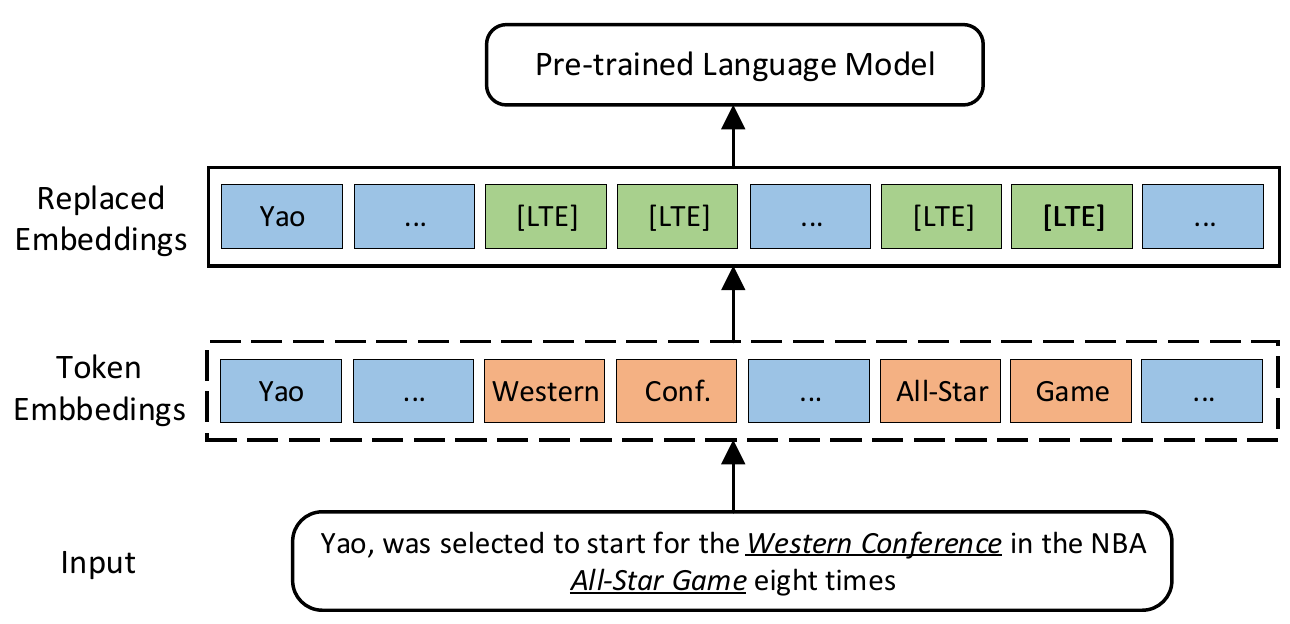}
\caption{Replacing the embeddings of long-tail entities with pseudo token embeddings.}
\label{figure7}
\end{figure}

DKPLM~\cite{b92} proposes the concept of long-tail entities which represent the entities not been fully learned by the model from the corpus. 
Strengthening the learning of such long-tail entities in the pre-training stage can enhance the model's understanding of semantic context and eventually the language representation. 
For this purpose, a measurement method KLT has been proposed to identify long-tail entities: the entities with a KLT score below the average in each sentence are regarded as the long-tail entities of that sentence. 
The KLT score of an entity $e$ is calculated as 
\begin{equation}
\label{eq2}
\begin{aligned}
KLT\left(e\right)=\mathbb{I}_{Freq\left(e\right)<R_{freq}}\cdot SI\left(e\right)\cdot KC\left(e\right),
\end{aligned}
\end{equation}
where the three terms in the equation represent the occurrence frequency of the entity in the corpus, the semantic importance, and the number of neighboring nodes within a certain number of hops in KG, respectively.
As illustrated in \cref{figure7}, DKPLM replaces the embeddings of long-tail entities detected in the text with pseudo token embedding as new input to the model.
For example, suppose that the input sentence is ``Yao, was selected to start for the Western Conference in the NBA All-Star Game eight times'' where ``Western Conference'' and ``All-Star Game'' have been identified as long-tail entities, so the embeddings of them will be replaced by pseudo token embeddings shown as ``[LTE]''. 
A pseudo token embedding is encoded by related triples in the knowledge graph and the entity's description in a certain way.
The $F_1$ scores of DKPLM~\cite{b92} on entity classification and relation classification are 2.1\% and 2.87\% higher than RoBERTa~\cite{b171} respectively, confirming that knowledge about long-tail entities could be incorporated into PLMs to obtain better language representation and higher model performance.

\subsubsection{Adding Knowledge Fusion Modules}
\label{sec:adding-knowledge-fusion-module}

\begin{figure}[!tb]
\centering
\includegraphics[width=1.0\columnwidth]{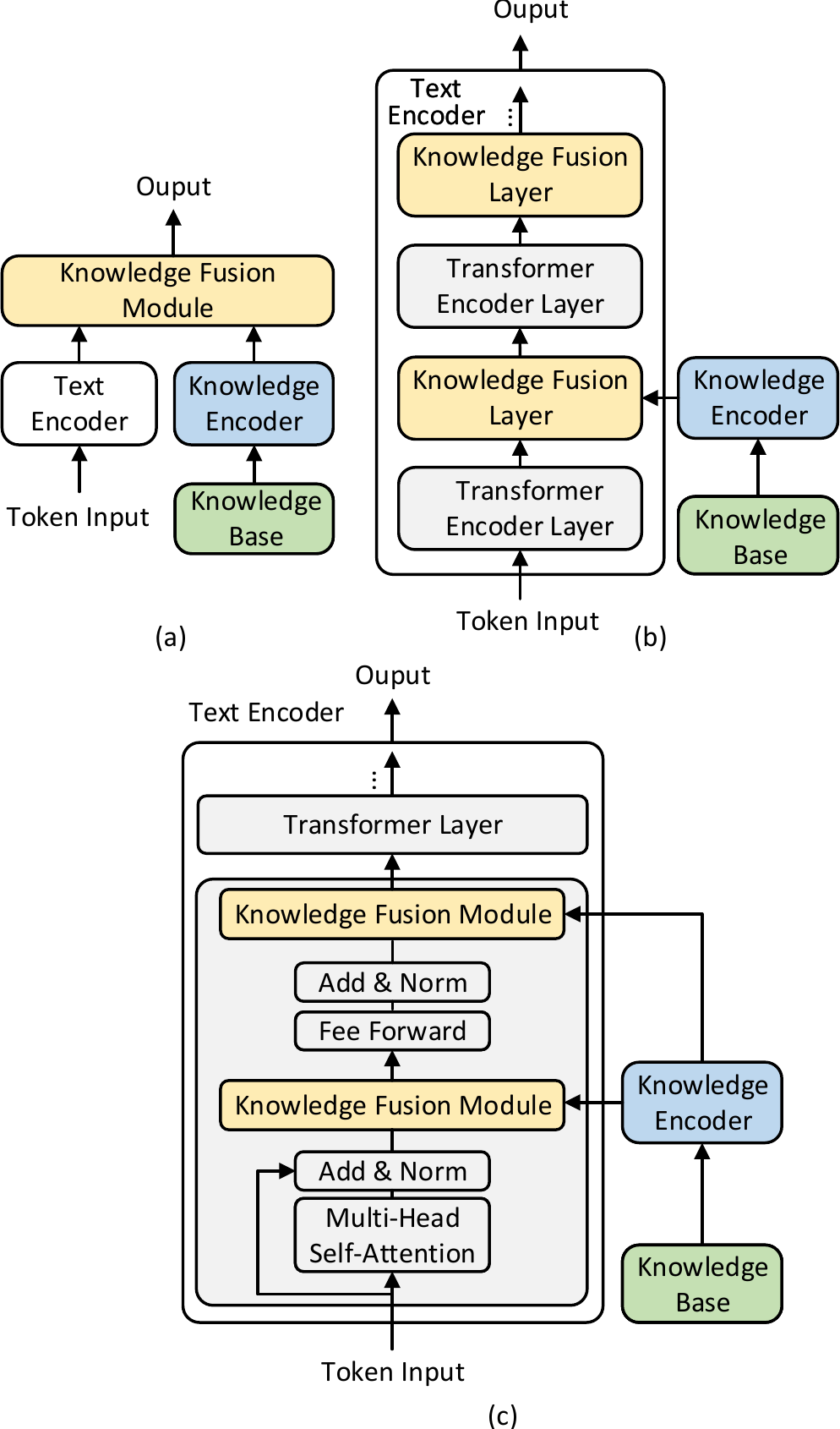}
\caption{Three different ways to add knowledge fusion modules to a PLM: (a) on top of the entire PLM, (b) between the Transformer layers of PLM, and (c) inside the Transformer layers of PLM.}
\label{figure8}
\end{figure}

Different from the methods introduced in \cref{sec:modifying-the-model-input}, the methods presented in this section all involve the fusion of different modal spaces. 
Specifically, the text and knowledge modalities are encoded differently, and additional modules are constructed for inter-modal fusion. 
As illustrated in \cref{figure8}, such knowledge fusion modules mainly appear in three positions:
\begin{enumerate}[(a)]
    \item on top of the entire PLM,
    \item between the Transformer layers of PLM,
    \item inside the Transformer layers of PLM.
\end{enumerate}

The method shown in \cref{figure8} (a) can be further divided into two categories. 
One is the T-K structure represented by ERNIE~\cite{b38} which mainly incorporates knowledge in the form of entity embeddings:
a T-Encoder is followed by a K- Encoder, where T- Encoder encodes the text corpus and K-Encoder integrates the entity embeddings in the knowledge space into the entity embeddings in the text space. 
Many KEPLMs follow this structure but differ in how they get entity embeddings.
The entity embedding in ERNIE is obtained by TransE which takes a single triple as a training sample and does not contain the information of that entity's neighbor nodes. 
Developed on top of this architecture, BERT-MK~\cite{b71} fully considers the information of neighbor nodes when learning the entity embedding in the knowledge space, incorporating more semantic information. 
CokeBERT~\cite{b83} found that the entity embeddings in the former method cannot change dynamically according to the textual context. 
To overcome this limitation, the closer the meaning of the neighbor node is to the text, the more its information will be incorporated into the entity embedding by CokeBERT.

\begin{figure}[!tb] 
\centering
\includegraphics[width=0.9\columnwidth]{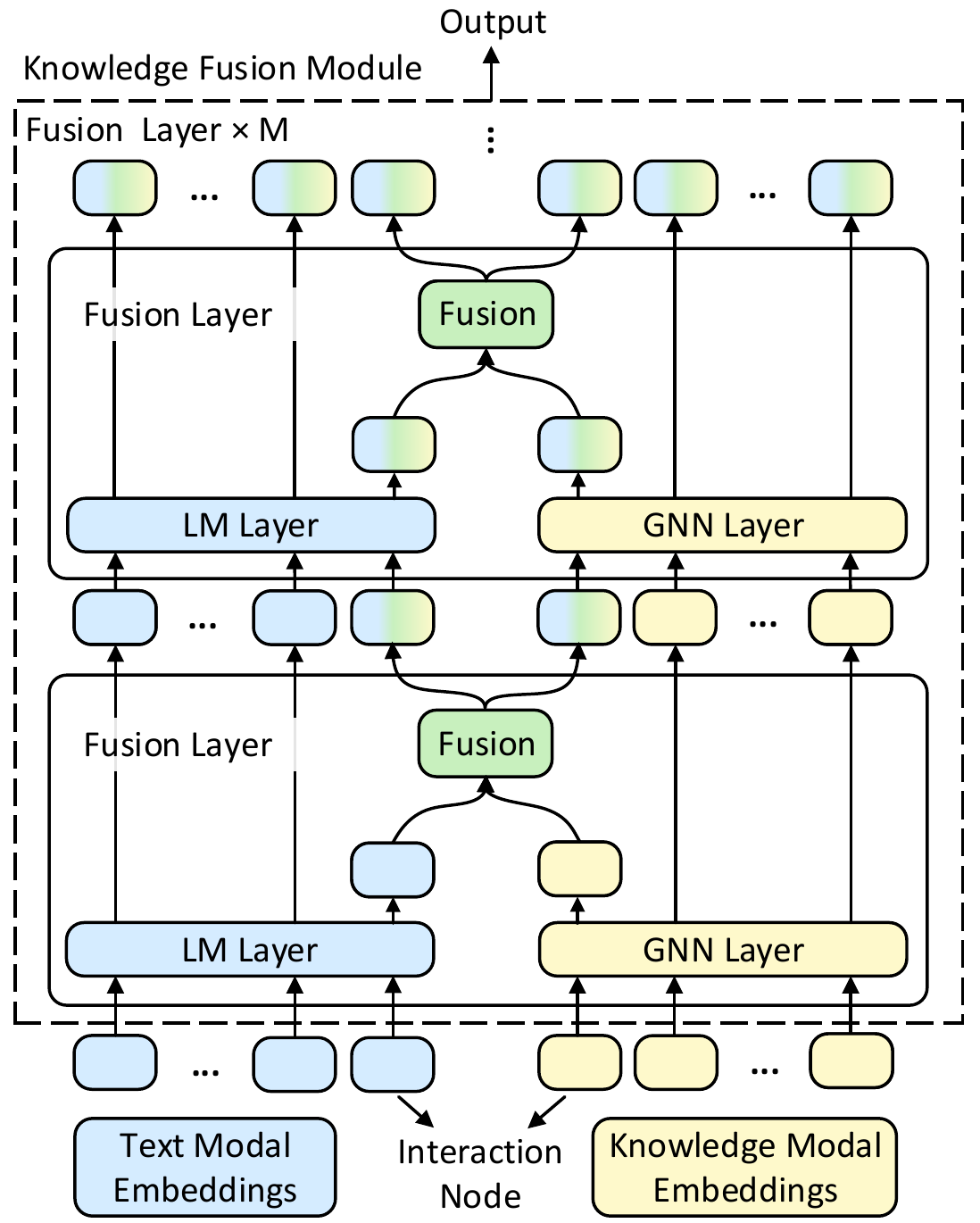}
\caption{The interaction nodes for text-knowledge information fusion~\cite{b53}.}
\label{figure9}
\end{figure}

The second class of methods attaches other knowledge fusion structures after the PLM. Some KEPLMs use the attention mechanism to fuse the information in the text and knowledge modalities. Kwon et al. exploited an attention mechanism to incorporate sentence-related triples into textual embedding representations~\cite{b94}. JointLK~\cite{b54} lets each question token attend on KG nodes and each KG node attend on question tokens, and the two modal representations fuse and update mutually by multi-step interactions. KET~\cite{b78} adopts a hierarchical self-attention mechanism to incorporate sentiment knowledge into text representations. Besides, Liu et al.~\cite{b86} encode the relevant triples in the context and then fused them with the embeddings of the text using a gate mechanism.
There are other works based on interaction nodes. Both modalities exchange information through interaction nodes. QA-GNN~\cite{b52} incorporated information from the text space into the knowledge space through interaction nodes and achieved good results in commonsense question answering. Inspired by this, GreaseLM~\cite{b53} set up interaction nodes in both modalities to learn the knowledge of that modality separately and then exchange information at the fusion layer to learn the knowledge of the other modality, as shown in \cref{figure9}.

The approach represented in \cref{figure8} (b) is to add a knowledge fusion module between the Transformer layers of PLM. 
KnowBERT~\cite{b95} adds new modules between Transformer Encoder blocks to incorporate knowledge about entities in sentences.
It considers the problem of polysemy that is ignored by ERNIE~\cite{b38}: for an entity that exhibits different semantic meanings in different contexts, the knowledge about it is incorporated according to its specific meaning. 
KG-BART~\cite{b57} added knowledge fusion modules between the Encoder and Decoder layers to integrate information from knowledge subgraphs into the textual representation through a multi-headed graph attention mechanism. 
JAKET~\cite{b96} divides the pre-trained language model into the first six layers and the last six layers. 
After the text passes through the first six layers of the encoder, the hidden layer representation is obtained, and so is the entity embedding representation. 
At each entity position in the text, the corresponding entity embedding representation is added and then input to the last six layers of the model for subsequent training. 
The knowledge space and the text space can cyclically reinforce each other for the learning of better representations.

The approach represented in \cref{figure8} (c) is to add a fusion module inside the Transformer Layer.
For example, KALA~\cite{b87} inserts the knowledge fusion module inside the Transformer block layer, which is inspired by the idea of modulation, i.e., to modulate the embeddings in the text space with the knowledge in the knowledge space.
Adding knowledge fusion modules in this way is intuitive, and the incorporated knowledge is mainly entity representation. 
Some methods consider the context of entities in the knowledge graph, e.g., BERT-MK~\cite{b71}; some others filter entity neighbor nodes for embedding based on text context, e.g., CokeBERT~\cite{b83}.

\subsubsection{Utilizing External Memory}
\label{sec:using-external-memory}

\begin{figure}[!tb] 
\centering
\includegraphics[width=0.9\columnwidth]{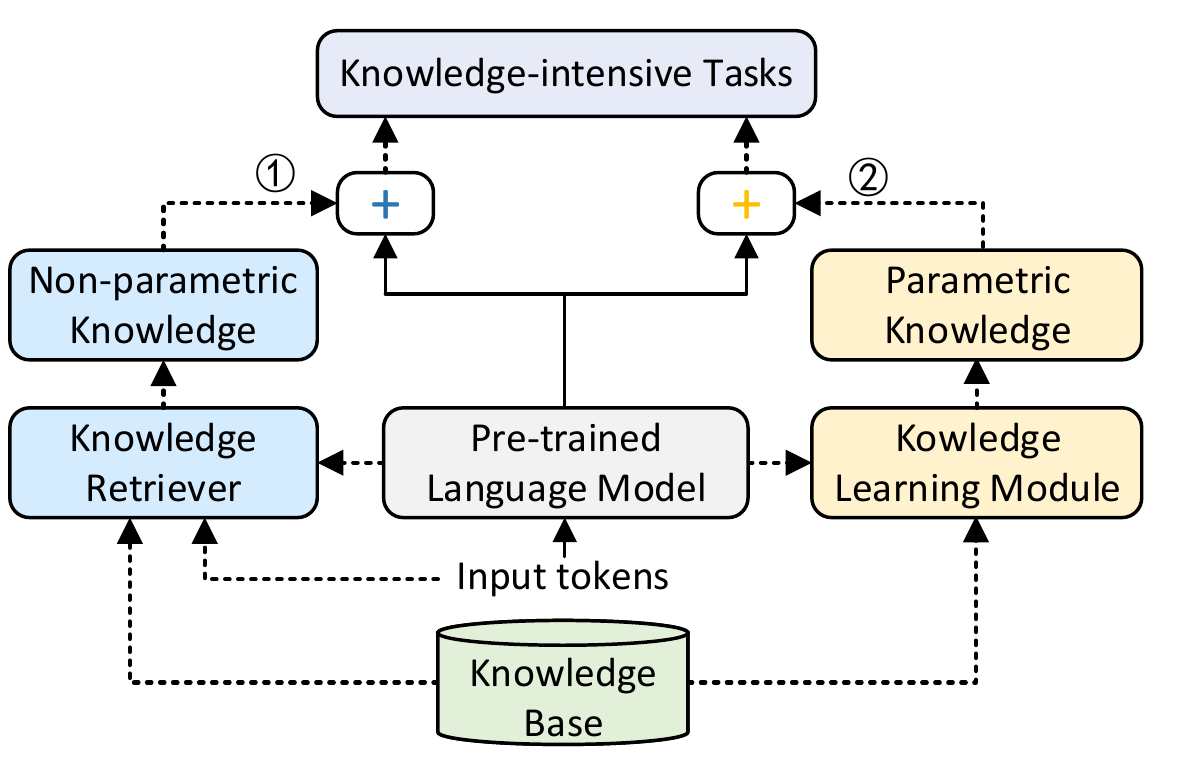}
\caption{Explicit incorporation of knowledge into PLMs via the utilization of external memory.}
\label{figure10}
\end{figure}

The third method for building KEPLMs explicitly uses external memory, and thus keeps the knowledge space and text space separate.
% , as illustrated in \cref{figure10}.

In \cref{figure10}, \ding{172} illustrates the method to apply non-parametric knowledge from external memory to downstream NLP tasks. 
KGLM~\cite{b97} selects and copies the facts from a related knowledge graph to generate factual sentences. 
In other words, it uses a knowledge base to expand the vocabulary to supply information it has never seen before. 
REALM~\cite{b98} introduces a knowledge retriever to help the model retrieve and process documents from the knowledge corpus, and thus improves the performance of open-domain question answering. 
It only needs to update the knowledge corpus if the world knowledge changes.

In \cref{figure10}, \ding{173} illustrates the method of learning parametric knowledge using an additional module independent of the PLM. 
K-Adapter~\cite{b30} adds adapters to learn parametric knowledge, and the parameters of the PLM itself remain unchanged during pre-training. 
Such adapters are independent of each other and can be trained in parallel. 
In addition, more adapters can be added when needed.

RAG~\cite{b99} that combines nonparametric and parametric memory outperforms other parametric-only and nonparametric-only models in three open domain question-answering tasks. 
Furthermore, for text generation, it can create more specific, diverse, and factual text than other parameter-only baselines. 
Wilmot et al.~\cite{b100} extend RAG~\cite{b99} by adding a memory module to improve the models' predictive performance.

When the knowledge base has undergone some changes, keeping the knowledge in external memory has the big advantage that the KEPLM does not require re-training, which is particularly helpful for the application domains where knowledge is updated frequently. 
% In contrast, the methods introduced in \cref{sec:modifying-the-model-input} and \cref{sec:adding-knowledge-fusion-module} if knowledge is incorporated in the pre-training phase, the model needs to be pre-trained again when the knowledge is updated.

\section{Evaluating KEPLMs}
\label{sec:Evaluating-KEPLMs}

This section presents methods for evaluating KEPLMs in terms of knowledge capacity, effectiveness, and efficiency. 

\subsection{Knowledge Capacity}

The amount of knowledge incorporated into KEPLMs could be assessed using \emph{knowledge probes} such as LAMA~\cite{b19} and LAMA-UHN~\cite{b162}. 
Intuitively, KEPLMs containing more knowledge would be more powerful for downstream NLP tasks. 

\subsubsection{LAMA}
LAnguage Model Analysis (LAMA) probe~\cite{b19} provides a series of completion statements that assess how much knowledge is stored in the model by the average accuracy of the model predictions. 
Knowledge sources for LAMA include Google-RE~\cite{b163}, T-Rex~\cite{b164}, ConceptNet~\cite{b49}, and SQuAD~\cite{b131}. 
The Google-RE corpus~\cite{b163} contains five kinds of relational triples, among which ``place of birth'', ``date of birth'', and ``place of death'' were selected by LAMA and transformed into fill-in-the-blank sentences according to the artificially constructed templates. 
For example, the triple ``place of birth'' is built as ``[S] was born in [O]'', where S represents the head entity and O represents the tail entity.
T-Rex~\cite{b164} is a subset of Wikidata containing 41 relations. 
The triples in it were also manually transformed into fill-in-the-blank sentences.
LAMA also selects triples from ConceptNet, covering 16 relations. 
For these triples, it finds the OMCS sentence containing both the head entity and the tail entity, then masks the tail entity within the sentence to construct a fill-in-the-blank sentence. 
LAMA~\cite{b19} selected 305 question-answer pairs from SQuAD and manually constructed fill-in-the-blank sentences. 
For example, the question ``Who developed the theory of relativity?'' was rewritten as "The theory of relativity was developed by $\_\_\_\_$”.
LAMA is generally recognized, and many existing work utilize LAMA~\cite{b19} to measure how much knowledge the model has learned, as shown in Table~\ref{tab:lama}.

% \usepackage{multirow}
\begin{table*}[!tb]
	\renewcommand{\arraystretch}{1.3}
	\centering
	\caption{Evaluation of Some KEPLMs on LAMA and LAMA-UHN Datasets}
	\label{tab:lama}
		\begin{tabular}{l|rrrr|rr}
			\hline\hline
			\multirow{2}{*}{\textbf{Model}} & \multicolumn{4}{c|}{\textbf{LAMA}} & \multicolumn{2}{c}{\textbf{LAMA-UHN}} \\
      \cline{2-7} 
			& \textbf{LAMA-Google-RE} & \textbf{LAMA-T-REx} & \textbf{ConceptNet} & \textbf{SQuAD} & \textbf{LAMA-UHN-Google-RE} & \textbf{LAMA-UHN-T-REx} \\
      \hline
			CoLAKE~\cite{b47} & {9.5} & {28.8} & {---}   & --- & {4.9}  & 20.4  \\
			KEPLER-Wiki~\cite{b46} & {7.3} & {24.6} & {18.7}  & 14.3      & {3.3}  & 16.5  \\
			KEPLER-W+W~\cite{b46} & {7.3} & {24.4} & {17.6}  & 10.8      & {4.1}  & 17.1  \\
			DKPLM~\cite{b92}  & {10.8}  & {32.0}   & {---}   & --- & {5.4}  & 22.9  \\
			KgPLM~\cite{b39}  & {9.2} & {27.9} & {---}   & --- & {4.9}  & 20.4  \\
 	    K-Adapter~\cite{b30}  & {7.0} & {29.1} & {---}   & --- & {3.7}  & 23.0  \\
			KALM~\cite{b90}   & {5.41}  & {28.12}    & {10.7}  & 11.89     & {---}  & --- \\
			EAE~\cite{b124}    & {9.4} & {37.4} & {10.7}  & 22.4      & {---}  & --- \\
			XLM-K~\cite{b27}  & {11.2}  & {29.7} & {15.7}  & 11.5      & {---}  & --- \\ 
      \hline\hline
		\end{tabular}
	% }
\end{table*}

In Table~\ref{tab:lama}, DKPLM~\cite{b92} performs better overall, which shows that long-tail entity-based learning helps the model remember factual knowledge. EAE~\cite{b124} learns entity representations directly from text rather than integrating entity knowledge into the model and performs well on all three datasets related to factual knowledge, which illustrates the effectiveness of the method.

\subsubsection{LAMA-UHN}

E-BERT~\cite{b162} found that for the fill-in-the-blank sentences of LAMA~\cite{b19}, the model may answer depending on the surface form of the entity name; for example, in predicting the language spoken by a person with an Italian-sounding name, the model would predict that the person speaks Italian.

To prevent the model obtaining answers from helpful entity names, E-BERT~\cite{b162} proposes LAMA-UHN (UnHelpfulNames), a subset of LAMA~\cite{b19} that focuses on factual knowledge, which deletes sentences with overly helpful entity names. 
Models such as CoLAKE~\cite{b47}, KEPLER~\cite{b46}, DKPLM~\cite{b92}, KgPLM~\cite{b39}, and K-Adapter~\cite{b30} are also evaluated on LAMA-UHN, as shown in Table~\ref{tab:lama}. 
The performances of the models on LAMA-UHN are much lower than those on LAMA, indicating that LAMA-UHN is more challenging to the model and it can better detect how much knowledge the model can actually learn.
%indicating->侧面说明

\subsubsection{Other Knowledge Probes}

In addition to LAMA and LAMA-UHN, there are other new knowledge probes. LPAQA~\cite{b165} considered that some sentences in LAMA and LAMA-UHN might be constructed inappropriately so that they only provide a lower bound estimate of the knowledge contained in an LM. That is, the model might know the answer but could not give the correct answer because of the inappropriate way of questioning. For example, for the sentence "Obama is a $\_\_\_\_$ by profession", the model is asked about Obama's profession, but the expression is unclear. If it is replaced by "Obama worked as a $\_\_\_\_$", it may predict more accurately. LPAQA aims to estimate the knowledge contained in LMs more accurately. Dolphs et al.~\cite{b166} applied example queries to LAMA probes, and the model's performance improved significantly, which also shows that we need to detect the knowledge contained in the language model properly to avoid underestimating the model. 
Unlike the above methods, AutoPrompt~\cite{b167} proposes an automated way to create prompts for measuring the amount of knowledge contained in LMs. This method saves time and effort. Moreover, prompts created by AutoPrompt can estimate knowledge in LM more accurately than manually created ones. 
In addition to general domain knowledge exploration, Meng et al. propose a biomedical knowledge exploration benchmark named MedLAMA~\cite{b168}.

\subsection{Effectiveness}
\label{sec:effectiveness}

We assess the effectiveness of a method by analyzing whether it maintains the original language representation capabilities and how many tasks' performances it can improve. 
We choose GLUE~\cite{b169} and KILT~\cite{b170} as corresponding benchmarks.

\subsubsection{General Language Understanding Tasks}

We choose the General Language Understanding Evaluation (GLUE) dataset~\cite{b169} as the benchmark to assess the general language representation capabilities maintained by KEPLMs.
It is a benchmark used to measure the performance of language models, containing nine natural language understanding tasks. 
It is the primary evaluation benchmark used by BERT.  
Many KEPLMs based on BERT or ROBERTa~\cite{b171} were tested on GLUE to explore whether incorporating knowledge affects the model's performance in natural language processing tasks. 
ERNIE~\cite{b38} has been tested on eight datasets of GLUE with essentially the same performance as BERT-base. 
It found that no additional knowledge is needed to process the tasks in GLUE, and the model does not cause a loss of textual information after incorporating knowledge. 
CoLAKE~\cite{b47}, SenseBERT~\cite{b32}, AMS~\cite{b59}, and other models have also been tested on GLUE and found that the way they incorporated knowledge did not affect the original language representation capabilities of the models. 
CoLAKE~\cite{b47} and AMS~\cite{b59} found that solving tasks in GLUE does not require encyclopedic and commonsense knowledge, respectively. 
Although we do not know whether all models incorporating knowledge affect the language representation capabilities of the models, the experiments done by the above models suggest that maintaining the original language representation capabilities of the model while incorporating knowledge is the goal pursued by the researchers.

\subsubsection{Knowledge-Intensive Language Tasks}

Most KEPLMs are designed for specific tasks, such as KG-BART~\cite{b57} for generative commonsense reasoning, ExBERT~\cite{b61} for natrual language inferance, GreaseLM~\cite{b53} for question answering, and so forth. 
These models may perform well in one task but fail in others. 
KEPLMs that can elevate performance on more tasks at the same time are of higher value, such as KGI~\cite{b172}, which can simultaneously improve the performance of fact checking, slot filling, open-domain QA, and dialog generation.

We choose the Knowledge Intensive Language Task (KILT) dataset~\cite{b170} as the benchmark to analyze the effectiveness of KEPLMs on different knowledge-intensive tasks. 
It contains 11 datasets in 5 categories of tasks, including Fact-checking, Entity linking, Slot filling, Open-domain QA, and Dialog generation. 
All tasks are based on the same Wikipedia snapshot, which aims to facilitate the development of general-purpose models and enable their comparisons.

\subsection{Efficiency}

We assess the efficiency of a KEPLM by considering its model size and required computational resources.

\subsubsection{Model Size}

Incorporating more knowledge would necessarily mean the expansion of the PLM. 
Usually the performance of a language model increases with its size (i.e., the number of the model parameters)~\cite{b26}, as shown in Table~\ref{tab:comparison_KEPLMs_T5}.
Briefly speaking, for the same level of performance, the smaller the model size, the more efficient the KEPLM.

\begin{table}[!tb]
  \renewcommand{\arraystretch}{1.3}
  \centering
  \caption{Comparison between KEPLMs and T5 in Terms of Model~Size and Task~Performance}
  \label{tab:comparison_KEPLMs_T5}
    \begin{tabular}{l| r| c| c}
      \hline\hline
      \multirow{2}*{\textbf{Model}} & \multirow{2}*{ \textbf{Params}} & \multicolumn{2}{c}{\textbf{Task Performance}}\\
      \cline{3-4}
      ~ & ~ &  \textbf{Natural Q.} &  \textbf{Web Q.} \\
      \hline
      T5-Base & 220M &  25.9 &  29.1\\
      T5-Large &  770M &  28.5 &  32.2\\
      T5-3B & 3B &  30.4 &  34.4\\
      T5-11B &  11B & 34.5 &  37.4\\
      \hline
      EAE~\cite{b124} & 367M &  ---  &39.0\\
      REALM~\cite{b98} &  330M &  40.4 &  40.7\\
      RAG-Token~\cite{b99} &  626M &  44.1 &  45.5\\
      RAG-Seq~\cite{b99}   &626M &  44.5 &  45.2\\
      \hline\hline 
  \end{tabular}
  % }
\end{table}

In Table~\ref{tab:comparison_KEPLMs_T5}, the performance of T5 on these two question-answering tasks increases with model parameters. 
That is to say, adding parameters can increase knowledge to some extent. 
However, the improvement of tasks is far less than the increment of parameters, which is expensive. 
KEPLMs such as EAE~\cite{b124}, REALM~\cite{b98}, and RAG~\cite{b99} adopt different methods to incorporate knowledge, which has far fewer parameters than T5-11B but better performance, indicating that incorporating knowledge properly can significantly improve performance efficiently.

\subsubsection{Computational Resources}

Computational resources usually increase with the model size and can also be a metric for evaluating the efficiency of KEPLMs. 
Precisely, we assess KEPLMs mainly through the training time and GPU or TPU used, as shown in Table~\ref{tab:metrics_KEPLMs}.

\begin{table}[!tb]
  \renewcommand{\arraystretch}{1.3}
  \centering
  \caption{Comparison of Computational Resourced Required by KEPLMs}
  \label{tab:metrics_KEPLMs}
  \setlength{\tabcolsep}{1.4mm}{%控制表总宽度
    \begin{tabular}{l|l|c|r}
      \hline\hline
      \textbf{Model} &  \textbf{GPU Type} & \textbf{GPU Num} & \textbf{Training}\\
      \hline
      SentiLARE~\cite{b28} & NVIDIA RTX 2080 Ti  &  4 & 20 hours \\
      DKPLM~\cite{b92}     & NVIDIA V100 16GB    &  8 & 12 hours \\
      CoLAKE~\cite{b47}    & NVIDIA V100 32GB    &  8 & 38 hours \\
      \hline\hline
  \end{tabular}}
\end{table}

We take Table~\ref{tab:metrics_KEPLMs} as an example to show how to compare the efficiency of KEPLMs by computing resources. 
First, we compare the GPUs used by the models. 
NVIDIA V100 GPU outperforms NVIDIA RTX 2080 Ti GPU. 
Then, we multiply the number of GPUs by the training time to roughly calculate the training time required for the model to run on just one GPU. 
SentiLARE takes 80 hours; DKPLM takes 96 hours; and CoLAKE takes 304 hours. 
The GPU used by SentiLARE is not as good as DKPLM and CoLAKE, and SentiLARE requires less training time, so SentiLARE requires the least computing resources. 
The GPU memory capacity used by DKPLM is smaller than CoLAKE, and the training time of DKPLM is smaller than CoLAKE, so DKPLM requires less computing resources than CoLAKE. 
Therefore, the computing resources required by these three models from low to high correspond to SentiLARE, DKPLM, and CoLAKE. 
The efficiency from high to low corresponds to SentiLARE, DKPLM, CoLAKE.

To sum up, we first look at the GPUs used by the models; then, we multiply the number of GPUs by the training time to roughly calculate the total training time using only one GPU. When GPU performance is close, the model with less total training time is more efficient.

\section{Applying KEPLMs}
\label{sec:Applying-KEPLMs}

KEPLMs are able to boost the performance of knowledge-intensive downstream tasks which can be grouped into two categories according to whether there is new natural language content created by the model. 

% and we classify these downstream tasks from two perspectives, i.e., model structures and application categories, aiming to provide researchers with two aspects of knowledge, namely, one is what kind of model structure is suitable for what types of tasks, and the other is what kind of downstream tasks can KEPLMs be applied to. 
% %从模型角度出发，什么样的模型适合什么类别的任务
% %根据任务本身划分
% The first classification method divides the downstream tasks of KEPLMs into \textit{Knowledge-Intensive Natural Language Understanding} (KINLU, in \cref{sec:KINLU}) and \textit{Knowledge-Intensive Natural Language Generation} (KINLG, in \cref{sec:KINLG}) according to whether the task uses Decoder or not; 
% and the second classification method categories them in accordance with their domains, introduced in \cref{sec:other-downstream-tasks}.

\subsection{Knowledge-Enhanced NLU}
\label{sec:KINLU}

KEPLMs based on Transformer encoder only or encoder-decoder could be used for \emph{natural language understanding} (NLU) tasks, such as entity typing, entity recognition, relationship extraction, sentiment analysis, question answering, language-based reasoning, and knowledge graph completion.

\subsubsection{Entity Typing}

Given an entity mention and its context, entity typing requires the model to classify the semantic type of the entity mention. 
FIGER~\cite{b101} and Open Entity~\cite{b102} are the most commonly used datasets. We found that models such as CoLAKE~\cite{b47}, KnowBERT~\cite{b95}, KEPLER~\cite{b46}, DKPLM~\cite{b92}, LUKE~\cite{b103} are only tested on Open Entity; 
ERICA~\cite{b40} is only tested on FIGER;
ERNIE~\cite{b38}, CokeBERT~\cite{b83}, and K-Adapter~\cite{b30} experiments on both datasets. 
Open Entity is more widely used, and we think the reasons are as follows. The training set of FIGER is annotated by remote supervision, and the testing set is manually annotated. Open Entity uses manual annotation for both datasets and has more types and finer-grained classification than FIGER. 
In addition, BERT-MK~\cite{b71} performs entity typing in the medical domain, using the datasets 2010 i2b2/VA~\cite{b104}, JNLPBA~\cite{b105}, and BC5CDR~\cite{b106}.

Most of the above methods insert special tokens before and after entity mentions in a given sentence to mark entity mentions (e.g., ``he had a differential diagnosis of [E] asystole [/E]'') and then use the embeddings of the special symbol preceding the entity mention (i.e., [E]) to predict the entity type.

Li et al.~\cite{b107} proposed a new dataset WikiWiki, containing 10 million Wikipedia articles with each entity connected to the knowledge graph of Wikidata~\cite{b17}. 
Compared with the existing fine-grained type recognition datasets, Wikiwiki is larger and more accurate, which can also be used for entity typing tasks.

\subsubsection{Entity Recognition}

The Named Entity Recognition (NER) task requires the model to identify the entity mentioned in a given text. 
In has been the basis for many NLP applications in both general and specific domains (like biomedical), as shown in \cref{tab:ner}.

\begin{table*}[!tb]
\renewcommand{\arraystretch}{1.3}
    \centering
    \caption{KEPLMs for Entity Recognition}
    \label{tab:ner}
    \begin{tabular}{p{2cm}|p{6cm}|p{9cm}}
    \hline\hline
    \textbf{Domain} &\textbf{Dataset} & \textbf{Model} \\
    \hline
    \multirow{2}{*}{General} & MSRA-NER~\cite{b108} & K-BERT~\cite{b45}, ERNIE~\cite{b89}, ERNIE~2.0~\cite{b109} \\
    \cline{2-3}
		~ & CoNLL-2003~\cite{b110} & KALA~\cite{b85}, LUKE~\cite{b103}, KMLMs~\cite{b111} \\
		\hline
		 \multirow{4}{*}{Biomedical} & English i2b2~\cite{b102,b112,b113} & UmlsBERT~\cite{b72}\\
     \cline{2-3}
		 ~ & DXY-NER~\cite{b114} & SMedBERT~\cite{b73}\\
     \cline{2-3}
		 ~ & Medicine$\_$NER~\cite{b115} & K-BERT~\cite{b45} \\
     \cline{2-3}
		 ~ & 
		 JNLPBA~\cite{b105}, BC5-chem $\&$ BC5-disease~\cite{b106}, NCBI-disease~\cite{b116}, BC2GM~\cite{b117} & KeBioLM~\cite{b74}, BioBERT~\cite{b118} \\
		 \hline
		 Finance & Finance$\_$NER~\cite{b119} & K-BERT~\cite{b45}\\
		 \hline
		 Social media &	WNUT-17~\cite{b120} &	KALA~\cite{b85}\\
		 \hline
		 Cross-lingual & WikiAnn NER~\cite{b121}  & 	KMLMs~\cite{b111} \\
		 \hline\hline
  \end{tabular}
  % }
\end{table*}

\subsubsection{Relation Extraction}

\begin{table*}[!tb]
\renewcommand{\arraystretch}{1.3}
    \centering
    \caption{KEPLMs for Relation Extraction}
    \label{tab:relation-extration}
    \begin{tabular}{p{2cm}|p{6cm}|p{9cm}}
    \hline\hline
    \textbf{Domain} & \textbf{Dataset} & \textbf{Model} \\
    \hline
    \multirow{2}*{General} 
    & TACRED~\cite{b122} & K-Adapter~\cite{b30}, ERNIE~\cite{b38}, ERICA~\cite{b40}, KEPLER~\cite{b46}, CokeBERT~\cite{b83}, LUKE~\cite{b103}, Glass et al.~\cite{b123}, DKPLM~\cite{b92}, EAE~\cite{b124} \\
    \cline{2-3}
    ~ & FewRel~\cite{b125} & ERNIE~\cite{b38}, KEPLER~\cite{b46}, CoLAKE~\cite{b47}, JAKET~\cite{b96},  CokeBERT~\cite{b83}\\
    \hline
    \multirow{3}*{Biomedical} 
    & 2010 i2b2/VA~\cite{b104}, GAD~\cite{b126}, EU-ADR~\cite{b127}  & BERT-MK~\cite{b71}\\
    \cline{2-3}
    ~ & DXY-NER~\cite{b114}, CHIP-RE~\cite{b128} & SMedBERT~\cite{b73}\\
    \cline{2-3}
    ~ & GAD~\cite{b126}, DDI~\cite{b129}, ChemProt~\cite{b130} & KeBioLM~\cite{b74} \\
    \hline\hline
    \end{tabular}
  % }
\end{table*}

KEPLMs could help to improve the extraction (and classification) of the relations between entities in a given text document.
In addition to the public field, this task is more commonly used in the biomedical domain.
As can be seen from \cref{tab:relation-extration}, the commonly used datasets in the general field are TACRED~\cite{b122} and FewRel~\cite{b125}, and TACRED is used more than FewRel. 
We find that all methods using these two datasets perform better on FewRel than on TACRED. 
That is to say, TACRED is more challenging than FewRel, so more and more methods tend to test model performance on TACRED. 
There are many work explicitly designed for the biomedical domain, and just like NER, relation classification tasks are helpful for models to learn domain-specific knowledge.

\subsubsection{Sentiment Analysis}

There are two kinds of sentiment analysis tasks: sentence-level sentiment analysis and aspect-level sentiment analysis. Sentence-level sentiment analysis requires models to determine the sentiment polarity of sentences, and commonly used datasets are Stanford Sentiment Treebank SST-2~\cite{b140} and Amazon-2~\cite{b141}. Aspect-level sentiment analysis requires the model to analyze the sentiment polarity in different aspects of the context, and commonly used datasets are SemEval-2014 Task 4~\cite{b142}. SentiLARE~\cite{b28} and SKEP~\cite{b79} obtained better results than pure PLM on both sentence-level and aspect-level tasks by incorporating sentiment knowledge. Through sentiment analysis, REMOTE~\cite{b80} could detect hate speech, and KET~\cite{b78} could detect sentiment in dialogues, which helps question-answering robots make better responses.

\subsubsection{Question Answering}
\label{sec:question-answering}

Question answering tasks include machine reading for question answering (MRQA), open-domain question answering (Open-domain QA), and multiple-choice question answering (Multiple-choice QA) according to the question form. We present commonly used datasets for each task in \cref{tab:qa}.

\begin{table*}[!tb]
	\renewcommand{\arraystretch}{1.3}
	\centering
	\caption{KEPLMs for Question Answering}
	\label{tab:qa}
	\begin{tabular}{p{2cm}|p{6cm}|p{9cm}}
		\hline\hline
		\textbf{Domain} & \textbf{Dataset} & \textbf{Model}\\
		\hline
		MRQA & SQuAD 1.1~\cite{b131}, NewsQA~\cite{b132}, TriviaQA~\cite{b133}, SearchQA~\cite{b134} & KT-NET~\cite{b31}, KgPLM~\cite{b39}\\
		\hline
		Open-Domain & Natural Questions~\cite{b135}, Web Questions~\cite{b136}, TriviaQA~\cite{b133}, SearchQA~\cite{b134} & REALM~\cite{b98}, K-ADAPTER~\cite{b29}, WKLM~\cite{b93}, and EAE~\cite{b124}  \\
		\hline
		Multi-Choice & CommonsenseQA~\cite{b137}, OpenBookQA~\cite{b138}, CosmosQA~\cite{b139}  & QA-GNN~\cite{b52}, GreaseLM~\cite{b53}, JointLK~\cite{b54}\\     
		\hline\hline
	\end{tabular}
	% }
\end{table*}

In \cref{tab:qa}, MRQA, also known as Extractive Question Answering, provides questions and related articles requires the model to find answers from the provided articles. The most commonly used dataset is SQuAD 1.1~\cite{b131}.

The Open-domain QA task gives the questions without the articles containing answers, requiring models to retrieve relevant articles. 
Methods such as REALM~\cite{b98}, K-ADAPTER~\cite{b29}, WKLM~\cite{b93}, and EAE~\cite{b124} are tested using some of the corresponding datasets in \cref{tab:qa}, and they achieve better results than the competitive baselines after incorporating encyclopedic knowledge.

The multiple-choice QA task requires the model to select the correct answer based on the question and the options given. The three datasets listed in \cref{tab:qa} are all commonsense question-answering tasks. Among them, CommonsenseQA~\cite{b137} has five options, OpenBookQA~\cite{b138} and CosmosQA~\cite{b139} have four options, and CommonsenseQA is the most widely used.

From \cref{tab:qa}, we can see that the datasets of Open-domain QA and MRQA overlap. Open-domain QA only removes the articles provided to the model based on MRQA and asks the model to retrieve them by itself so that they can share datasets.

\subsubsection{Language-based Reasoning}

Representative KEPLMs used for reasoning tasks include
SMedBERT~\cite{b73} and Li et al.~\cite{b159} for \emph{natural language inference},
KMLMs~\cite{b111} for logical reasoning, 
Andor et al.~\cite{b161} for mathematical reasoning,
Chang et al.~\cite{b64} and Vokenization~\cite{b82} for commonsense reasoning,
CoCoLM~\cite{b70} for reasoning about the temporal order of events, and 
VALM~\cite{b81} for reasoning about object color and size.

\subsubsection{Knowledge Graph Completion}

Knowledge graphs often suffer from incompleteness, and many relationships between entities are missing. 
KEPLMs can help infer missing links and complement knowledge graphs to a certain degree.

Models such as GLM~\cite{b55} were tested on the WN18RR~\cite{b150} and CKBC~\cite{b151} sets. It outperforms some translation-based graph embedding models and graphs convolutional networks on WN18RR. CKBC is a generic knowledge graph derived from OMCS~\cite{b152}, and GLM~\cite{b55} outperforms KG-BERT~\cite{b153}, a model specifically designed for the knowledge graph completion task, on this dataset.

K-PLUG~\cite{b77} performs the e-commerce knowledge graph completion task on MEPAVE~\cite{b154}, which gives a textual description of a product and asks the model to output the attribute values of the product.

\subsection{Knowledge-Enhanced NLG}
\label{sec:KINLG}

KEPLMs based on Transformer decoder only or Transformer encoder-decoder could be used for \emph{natural language generation} (NLG) tasks, such as sentence generation, dialogue generation, question generation, and answer generation.

\subsubsection{Sentence Generation}

The sentence generation task requires models to generate reasonable sentences, and commonly used datasets are CommonGen~\cite{b143} and ROCStories~\cite{b144}. CommonGen requires models to generate a coherent, proper sentence based on 3-5 given concepts, while KG-BART~\cite{b57} does so by incorporating commonsense knowledge subgraphs. Models such as GRF~\cite{b60}, Guan et al.~\cite{b145}, and Guan et al.~\cite{b63} can generate plausible story endings with the help of commonsense knowledge.

\subsubsection{Dialogue Generation}

Dialogue generation tasks require the model to generate responses based on the context of the dialogue. KnowledGPT~\cite{b146} chose to conduct experiments on Wizard~\cite{b147} and CMU$\_$DoG~\cite{b148} datasets. Wizard has a wide range of topics, while CMU$\_$DoG only focuses on the movie domain.

\subsubsection{Question Generation}

The question generation task requires the model to generate questions based on the answers. RAG~\cite{b99} proposes the Jeopardy Question Generation task, where Jeopardy consists of trying to guess an entity from the facts about it. For example, given the answer ``The World Cup'', models need to generate relevant fact that points to the answer, like ``In 1986, Mexico scored as the first country to host this international sports competition twice.''

\subsubsection{Answer Generation}

Unlike standard question-answering (QA) tasks mentioned above in \cref{sec:question-answering}, open-domain \emph{abstractive} QA, aka zero-shot QA or closed-book QA, requires the model to generate answers by itself rather than finding answers from passages or selecting from options. 
RAG~\cite{b99} only uses questions and answers from the dataset of MSMARCO NLGT task v2.1~\cite{b149}, treating it as the open-domain abstractive QA task and outperforming the baseline model BART~\cite{b15}.

\section{Future Directions}
\label{sec:challenges-future}

In the above sections, we have presented KEPLMs from multiple perspectives, but there are still some other opportunities. 
Here we outline and discuss a few promising research directions for KEPLMs.

\textit{\underline{Utilizing More Types of Knowledge.}}
As mentioned in \cref{sec:knowledge-sources}, existing KEPLMs have considered many types of knowledge, but there are other types of knowledge worth investigating.
For example, temporal knowledge graphs such as HyTE~\cite{b173} contain events that reflect the relationships between different entities over time, so incorporating them into PLMs could help to perform time-related reasoning tasks.
Moreover, the phenomenal success of ChatGPT has demonstrated the power of incorporating the knowledge of human intentions and preferences into PLMs directly through Reinforcement Learning from Human Feedback (RLHF)~\cite{ouyang2022training}.

\textit{\underline{Improving the Effectiveness of Knowledge Incorporation.}}
As described in \cref{sec:Building-KEPLMs}, a variety of technical approaches to incorporating knowledge into PLMs have been proposed.
Some of those methods such as KEPLER~\cite{b46} and CokeBERT~\cite{b83} rely on sophisticated joint pre-training of PLMs and KG embeddings. 
However, KEPLER~\cite{b46} performs worse on entity typing and relation classification tasks than LUKE~\cite{b103} which contains only entity-level knowledge. 
CokeBERT~\cite{b83} is slightly better than LUKE~\cite{b103} on some datasets, but it is not as efficient as LUKE~\cite{b103}. 
% There still seems to be much room for more effective incorporation of knowledge.
% 
Hou et al.~\cite{b177} proposed the Graph Convolution Simulator to detect knowledge integrated into PLMs. 
Their examination of ERNIE~\cite{b38} and K-Adapter~\cite{b30} revealed that those KEPLMS have only incorporated a small amount of factual knowledge.
There still seems to be much room for more effective incorporation of knowledge into PLMs.

\textit{\underline{Improving the Efficiency of Knowledge Incorporation.}}
Most existing work about KEPLMs only report improvements with respect to model performance, and only a few assess the costs of knowledge incorporation as well. 
More time-efficient and space-efficient solutions to KEPLMs are desired.
Many methods, such as CoLAKE~\cite{b47} and ERNIE~\cite{b38}, carry out knowledge incorporation in the pre-training stage, while some others like K-BERT~\cite{b45}, K-Adapter~\cite{b29}, and Syntax-BERT~\cite{b29} carry out knowledge incorporation in the fine-tuning stage. 
The time cost of knowledge incorporation in the pre-training stage is greater than doing that in the fine-tuning stage. 
It deserves more investigation to minimize the overhead in the pre-training stage while maintaining good performance. 
Besides, knowledge incorporation may also increase the inference overhead of the model. 
For example, GRF~\cite{b60} and KG-BART~\cite{b57} involve the construction of knowledge sub-graphs, which makes their inference time much longer. 
More efficient inference strategies need to be developed for KEPLMs to facilitate their practical applications.
The additional space consumption of KEPLMs must also be carefully considered before their deployment. 
For example, FaE~\cite{b156} needs an external entity memory and a factual memory containing millions of knowledge triples. 
RAG~\cite{b99} relies on a non-parametric knowledge corpus containing tens of millions of documents. 
Not all of these integrated entities or facts are equally useful: some of them probably play more important roles than others in enhancing the PLM. 
Therefore, selecting and storing only the most critical subset of knowledge may significantly reduce the space overhead with a small sacrifice in performance.
In addition to avoiding the incorporation of less important knowledge, model compression techniques~\cite{b179} can be used to reduce the computational overhead of KEPLMs. 
For example, quantization~\cite{b180}, knowledge distillation~\cite{b181}, and parameter sharing~\cite{b182} can all be applied to KEPLMs to improve their time and space efficiency.

\textit{\underline{Exploring Other Knowledge-Intensive Tasks.}}
In addition to the downstream NLP tasks listed in \cref{sec:Applying-KEPLMs}, some other less-explored applications may also benefit from KEPLMs.
For example, KEPLMs are likely to improve the correctness of machine translation and the factualness of text summarization~\cite{b178}. 

\textit{\underline{Building A Unified KEPLM for Multiple Tasks.}}
Most of the existing KEPLMs are designed for specific knowledge-intensive NLP tasks.
Currently to achieve SOTA performance for different tasks, one often needs to train a different KEPLM for each of them. 
It is desirable to develop a unified KEPLM for multiple tasks so as to avoid the costly proliferation of KEPLMs. 
There have been some early attempts towards this direction, such as KGI~\cite{b172} which is trained to improve the performance on four different tasks in the KILT~\cite{b170} benchmark.
% 
% New training methods, such as prompt learning, are also helpful to get unified models. 
% Instead of adapting pre-trained LMs to downstream tasks done by common PLMs, it adapts downstream tasks to PLMs, i.e., designing different prompts for different downstream tasks to guide the model to recall relevant knowledge, which is worthy of further explorations.
% 
%Correspondingly,  evaluation methods for unified KEPLMs such as KILT~\cite{b170} are worth exploring, which can promote the development of unified KEPLMs.

\textit{\underline{Performing Zero/Few-shot Learning.}} 
In some application domains, there are little quality labelled data, therefore zero-shot learning or few-shot learning will be particularly useful. 
Thanks to the knowledge built into KEPLMs, they are more able than standard PLMs to overcome the data scarcity problem and tackle many zero/few-shot learning tasks. 
KALM~\cite{b90} signals the existence of entities to the input in pre-training to integrate knowledge, significantly improving zero-shot question-answering tasks. 
Li et al.~\cite{b107} introduced fine-grained type knowledge of entities, achieving superior performance in zero-shot dialog state tracking. 
Other than incorporating knowledge into PLMs, one can also exploit external knowledge in prompt engineering~~\cite{Shengding-Prompt-tuning-ACL-2022,Xiang-KnowPrompt-WWW-2022,Hongbin-OntologyPrompt-WWW-2022,Brate-KGPrompts-ISWC-2022}. 
It would be interesting to investigate how to maximize the combined effect of knowledge-enhanced PLMs and knowledge-enhanced prompt engineering together.

\textit{\underline{Achieving Better Interpretability and Robustness.}}
The interpretability of a model measures how easily a human can understand its results and predictions.
Schuff et al.~\cite{b183} investigated whether incorporating external knowledge can help to explain natural language inference tasks. 
They have argued that there is a discrepancy between the automatic evaluation method of models and manual scoring, and the effectiveness of automatic evaluation needs to be reconsidered. 
Akyürek et al.~\cite{b184} attempted to trace the predictions made by the model back to training data. 
Cao et al.~\cite{b175} and LEFA~\cite{b176}, on the other hand, attempted to locate the knowledge stored in the model. 
% There is still much potential to improve the interpretability of the model in the future.
% 
The robustness of a model refers to its resistance to input disturbances or adversarial attacks etc. 
Li et al.~\cite{b159} improved the robustness of the model by introducing external lexical knowledge into the attention mechanisms. 
Glass et al.~\cite{b123} demonstrated adaptive capabilities on new datasets to illustrate the model's robustness.
% The above work have been explored in this area that can provide inspiration for robustness. 
There are not many existing studies which try to improve the interpretability or robustness of KEPLMs. 
More in-depth investigations on these aspects would be helpful.  

\section{Conclusion}\label{sec:conclusion}

In summary, this survey provides a comprehensive view of current advances in the rapidly evolving field of KEPLMs.
We begin by briefly introducing KEPLMs and describing the knowledge types/formats along with the methods for knowledge incorporation. 
PLMs can be enhanced by a wide range of knowledge, with encyclopedic and commonsense knowledge being most widely used, and domain-specific knowledge being increasingly explored. 
Different types of knowledge come in various forms, e.g., knowledge graphs could be integrated into PLMs directly as triples or indirectly through embeddings. 
The methods for knowledge incorporation can be classified into two main categories, implicit and explicit. 
Implicit incorporation does not put external knowledge into the model but employs knowledge-guided masking strategies or knowledge-related pre-training tasks to mine and learn knowledge from the pre-traineing corpus.
Explicit incorporation can be adding knowledge to the input, integrating knowledge through fusion structures, or storing knowledge in external memory and retrieving it when needed. 
We then introduce some off-the-shelf methods for assessing the effectiveness of KEPLMs by detecting the amount of knowledge learned by the model, propose metrics for assessing model efficiency, and suggest assessing the generality of the model based on whether it can simultaneously boost performance on various tasks. 
After that, we present a list of knowledge-intensive tasks and some application areas worth considering. 
A final discussion of KEPLM research directions concludes this paper. We hope this will inspire researchers to explore KEPLMs further in the future.
Finally, we discuss potential research directions for KEPLMs, which we hope will inspire future research in this area.

\ifCLASSOPTIONcompsoc
  \section*{Acknowledgments}
  Yong Chen is supported by the Young Scientists Fund of the National Natural Science Foundation of China (Grant No. 62006005), and the National Key Research and Development Program of China (No. SQ2022YFC3300043).
\else
  \section*{Acknowledgment}
  The authors would like to thank...
\fi
%\fi

\ifCLASSOPTIONcaptionsoff
  \newpage
\fi



%\bibliographystyle{IEEEtran}

\bibliographystyle{IEEEtran}
\bibliography{references}



% \end{thebibliography}


% 作者介绍
\begin{IEEEbiography}[{\includegraphics[width=1in,height=1.25in,clip,keepaspectratio]{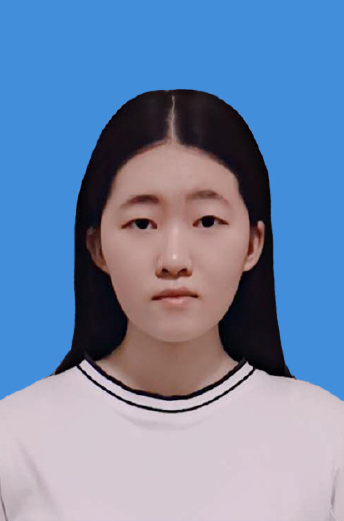}}]{Chaoqi Zhen}
	received the bachelor's degreein computer science and technology from theBeijing Unibersityof Posts and Telecommuni-cations, Beijing, China, in 2020. She is current-ly working towrd the master's degree in com-puter science and technology from the BeijingUnibersity of Posts and Telecommunications,Beijing, China. Her research interests lie innatural language processingandknowledgerepresentation learning.
\end{IEEEbiography}

\begin{IEEEbiography}[{\includegraphics[width=1in,height=1.25in,clip,keepaspectratio]{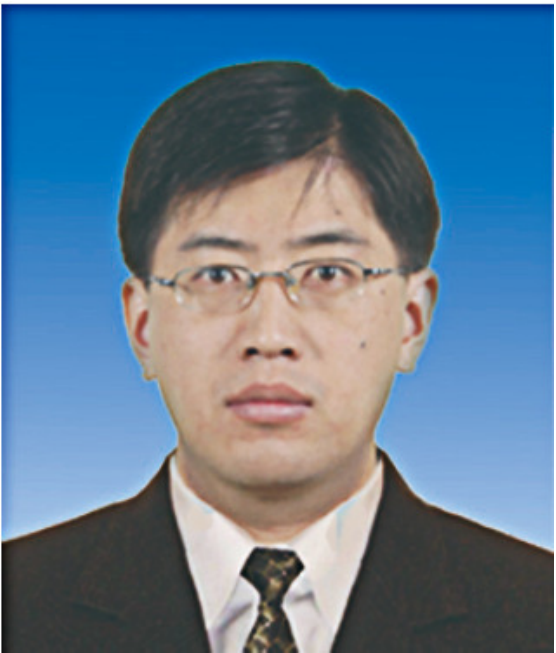}}]{Yanlei Shang}
	received the PhD degree in computer science and Technology from Beijing Unibersity of Posts and Telecommunications, Beijing, China, in 2006. He is currently working as a professor with the school of computer science, Beijing Unibersity of Posts and Telecommunica-tions, Beijing, China. His research interests mainly include cloud computing, big data storage and analytics, artificial intelligence and deep learning.
\end{IEEEbiography}

%\newpage


%\vfill

\begin{IEEEbiography}[{\includegraphics[width=1in,height=1.25in,clip,keepaspectratio]{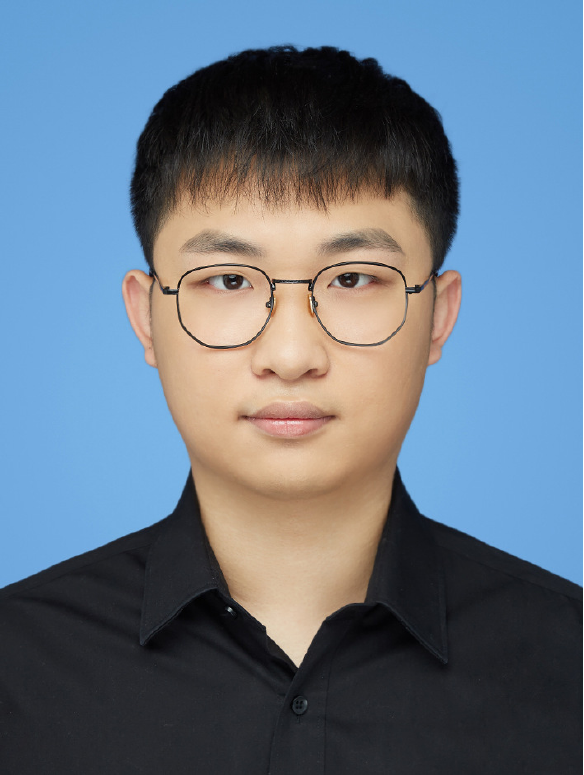}}]{Xiangyu Liu}
	received the bachelor's degree in computer science and technology from the Beijing Unibersity of Posts and Telecom-munications, Beijing, China, in 2020. She is currently working towrd the master's degree in computer science and technology from the Beijing Unibersity of Posts and Telecommunications, Beijing, China. His research interests lie in natural language processing, and knowledge representation learning.
\end{IEEEbiography}

\begin{IEEEbiography}[{\includegraphics[width=1in,height=1.25in,clip,keepaspectratio]{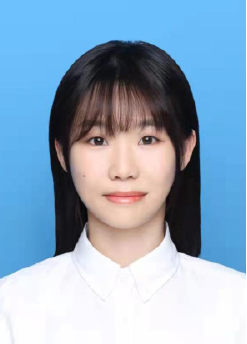}}]{Yifei Li}
	received the bachelor's degree in computer science and technology from the Beijing Unibersity of Posts and Telecom-munications, Bei-jing, China, in 2020. She is currently working towrd the master's degree in computer science and technology from the Beijing Uni-bersity of Posts and Telecommunications, Beijing, China. Her re-search interests lie in natural language processing, and knowledge acquisition.
\end{IEEEbiography}

\begin{IEEEbiography}[{\includegraphics[width=1in,height=2in,clip,keepaspectratio]{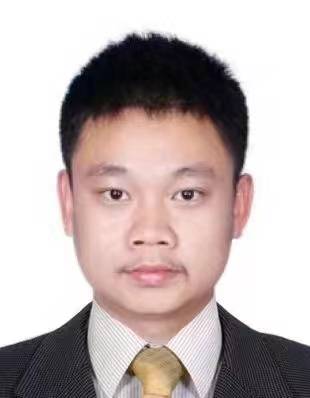}}]{Yong Chen}
	is now an associate professor in school of computer science, Beijing University of Posts and Telecommunications, Beijing, China. He received the Ph.D. degree in computer science and engineering from Beihang University (BUAA), Beijing, China, in 2019; and worked as a “Boya” postdoctoral with the Key Lab of Machine Perception, School of Electronics Engineering and Computer Science, Peking University, Beijing, China, from 2019 to 2021. He has been funded as a visiting PhD student at Birkbeck and UCL from January 2018 to January 2019. His research interests include machine learning, data mining, big data, numerical optimization and interpretable differential calculation.
\end{IEEEbiography}

\begin{IEEEbiography}[{\includegraphics[width=1in,height=2.5in,clip,keepaspectratio]{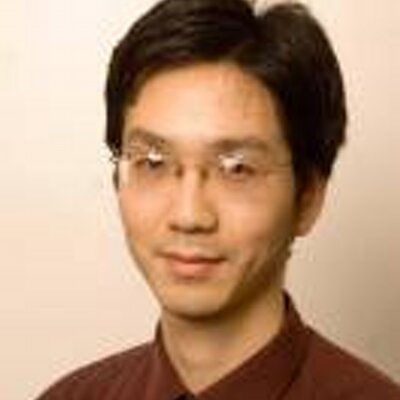}}]{Dell Zhang}
currently leads the Applied Research team at Thomson Reuters Labs in London, UK.
Prior to this role, he was a Tech Lead Manager at ByteDance AI Lab and TikTok UK, a Staff Research Scientist at Blue Prism AI Labs, and a Reader in Computer Science at Birkbeck College, University of London. 
He is a Senior Member of ACM, a Senior Member of IEEE, and a Fellow of RSS. 
He got his PhD from the Southeast University (SEU) in Nanjing, China, and then worked as a Research Fellow at the Singapore-MIT Alliance (SMA) until he moved to the UK in 2005. His main research interests include Machine Learning, Information Retrieval, and Natural Language Processing. 
He has published 110+ papers, graduated 11 PhD students, received multiple best paper awards, and won several prizes from international data science competitions. 
\end{IEEEbiography}


\end{document}